%% file: arslan.arxiv.final.tex
\documentclass{llncs}
\pdfoutput=1

\usepackage{makeidx}  

\usepackage{times}
\usepackage[dvips]{graphicx}\usepackage{graphicx}
\usepackage{amsmath}
\usepackage{amssymb}
\usepackage{subfigure}
\usepackage{wrapfig}
\usepackage{epsfig}
\usepackage{comment}
\usepackage{xspace}
\usepackage{cite}
\usepackage{verbatim}
\usepackage{color}

\usepackage{mathtools}

\newtheorem{defn}{\bf{Definition}}

\newcommand{\PISq}{PI$^2$\xspace}

\newcommand{\vx}{{\bf x}}
\newcommand{\xx}{\mathbf{x}(\cdot)}

\newcommand{\vz}{{\bf z}}

\newcommand{\vf}{{\bf f}}

\newcommand{\vG}{{\bf G}}
\newcommand{\vu}{{\bf u}}

\newcommand{\vI}{{\bf I}}

\newcommand{\vw}{{\bf w}}

\newcommand{\vE}{{\bf E}}

\newcommand{\vF}{{\bf F}}

\newcommand{\vV}{{\bf V}}
\newcommand{\vA}{{\bf A}}

\newcommand{\vC}{{\bf C}}

\newcommand{\vB}{{\bf B}}

\newcommand{\vsigma}{{\mbox{\boldmath$\sigma$}}}

\newcommand{\VSigma}{{\mbox{\boldmath$\mathrm\Sigma$}}}

\newcommand{\vcalE}{\ensuremath{\mathcal{E}}}
\newcommand{\vcalF}{\ensuremath{\mathcal{F}}}

\newcommand{\vcalJ}{\ensuremath{\mathcal{J}}}

\newcommand{\vmesP}{\ensuremath{\mathsf{p}}}
\newcommand{\vmesQ}{\ensuremath{\mathsf{q}}}

\newcommand{\Real}{\ensuremath{\mathbb{R}}}

\newcommand{\half}{\mbox{$\frac{1}{2}$}}
\newcommand{\onerho}{\mbox{$\frac{1}{\rho}$}}

\usepackage[vlined,linesnumbered,boxruled]{algorithm2e}

\input{commands_integrated.tex}

\newcommand\myfootnote[1]{%
  \begingroup
  \renewcommand\thefootnote{}\footnote{#1}%
  \addtocounter{footnote}{-1}%
  \endgroup
}

\begin{document}
%
%
\pagestyle{headings}  
\addtocmark{Stochastic Control} 
%
%
%
\mainmatter              
\title{Information-Theoretic Stochastic Optimal Control \\via Incremental Sampling-based Algorithms}
\titlerunning{Stochastic Control}  
\author{Oktay Arslan \quad Evangelos Theodorou \quad Panagiotis Tsiotras\myfootnote{The authors are with the Daniel Guggenheim School of Aerospace Engineering and the Institute for Robotics and Intelligent Machines, Georgia Institute of Technology, Atlanta, GA.}}

\authorrunning{Oktay Arslan et al.} 
%
\tocauthor{Oktay Arslan, Evangelos Theodorou, and Panagiotis Tsiotras}

\institute{}

\maketitle              

\begin{abstract}
This paper considers optimal control of dynamical systems which are represented by nonlinear stochastic differential equations. It is well-known that the optimal control policy for this problem can be obtained as a function of a value function that satisfies a nonlinear partial differential equation, namely, the Hamilton-Jacobi-Bellman equation. This nonlinear PDE must be solved backwards in time, and this computation is intractable for large scale systems. Under certain assumptions, and after applying a logarithmic transformation, an alternative characterization of the optimal policy can be given in terms of a path integral. Path Integral (PI) based control methods have recently been shown to provide elegant solutions to a broad class of stochastic optimal control problems. One of the implementation challenges with this formalism is the computation of the expectation of a cost functional over the trajectories of the unforced dynamics. Computing such expectation over trajectories that are sampled uniformly may induce numerical instabilities due to the exponentiation of the cost. Therefore, sampling of low-cost trajectories is essential for the practical implementation of PI-based methods. In this paper, we use incremental sampling-based algorithms to sample useful trajectories from the unforced system dynamics, and make a novel connection between Rapidly-exploring Random Trees (RRTs) and information-theoretic stochastic optimal control. We show the results from the numerical implementation of the proposed approach to several examples.

\end{abstract}

\section{Introduction}

In \cite{TheodorouYale2013, TheodorouCDC2013}, the authors showed the connection between Kullback-Leibler (KL) and Path Integral (PI) control with an information-theoretic view of stochastic optimal control. In addition, the authors derived the iterative  path integral optimal control without relying on policy parameterizations, as in \cite{theodorou2010}. We review the work in \cite{TheodorouYale2013, TheodorouCDC2013} starting with the definitions of free energy and relative entropy and their  connections to dynamic programming. In addition, we discuss how the iterative scheme developed in \cite{TheodorouYale2013} and \cite{TheodorouCDC2013} can be modified to incorporate incremental sampling-based methods such as Rapidly-exploring Random Trees (RRTs) to guide sampling.

Within  the mathematical framework of path integral control, the Feynman-Kac lemma plays an essential role, since it creates a connection between Stochastic Differential Equations (SDEs) and backward Partial Differential Equations (PDEs).  This fundamental connection between SDEs and backward PDEs has inspired new avenues for the  development of  stochastic control algorithms such as Policy Improvement with Path Integrals (\PISq) \cite{theodorou10b}  that rely on forward sampling.
\PISq   has been applied to a plethora of motor control tasks from robotic object manipulation  and  locomotion to general  trajectory optimization and gain scheduling\cite{Buchli_IJRR_2011,Freek2012_SequenceManipulation,theodorou10b,MorimotoPathIntegral}, but it relies on a suitable parameterization of the optimal control policy.   While policy parameterization such as Dynamic Movement Primitives (DMPs)~\cite{ijspeert2013} improves sampling by steering trajectories in high-dimensional state spaces towards areas of interest,  it  does not exploit the feedback structure provided by the path integral control framework.  In  \PISq
trajectories are sampled from the initial state of the task, the optimal parameter variations are computed, and the parameters are updated. In the next iteration, trajectories are sampled again from the same initial state and the iterative process continues until convergence.  It is clear that  in the case of policy parameterization  one has to explicitly design the structure of the feedback control policy and then treat the gains as parameters to be optimized.

In this work, we use an alternative approach, which steers state trajectories towards relevant areas of the state space without the requirement of policy parameterization.  In addition,  the proposed approach  improves sampling, while also allowing the use of path integral control in a feedback form.


\section{Notation}

A \textit{probability space} is a triple ($\mathrm{\Omega}$, $\mathcal{F}$, $\vmesP$) where $( \mathrm{\Omega}, \vcalF) $ is a measurable space with
$\mathrm{\Omega}$ a non-empty set, which is called the \textit{sample space}, $\mathcal{F} \subseteq 2^{\mathrm{\Omega}}$ a $\sigma$-algebra of subsets of $\mathrm{\Omega}$, whose elements are called events, and $\vmesP$ is a \textit{probability measure} on $\mathcal{F}$, that is, $\vmesP$ is a finite measure on $\mathcal{F}$ with $\vmesP (\mathrm{\Omega}) = 1$.

A real random variable is a function $X : \mathrm{\Omega} \rightarrow \mathbb{R}$ with the property that $\{ \omega \in \mathrm{\Omega} : X(\omega) \leq x \} \in \mathcal{F}$ for each $x \in \mathbb{R}$. Such a function is said to be $\mathcal{F}$-measurable. An extended (real) random variable can also take the values $\pm \infty$.
If $X$ is a random variable on the probability space ($\mathrm{\Omega}$, $\mathcal{F}$, $\vmesP$), then its \textit{expectation} is defined by
\begin{equation} \label{eq:defEXP}
	\mathbb{E}_{\vmesP} \left[ X \right] = \int_{ \mathrm{\Omega} } X (\omega) \, \dif{\vmesP(\omega) },
\end{equation}
provided that the integral in the right-hand side exists. As usual, and for notational simplicity, in the sequel we will drop the explicit dependence on $\omega \in \mathrm{\Omega} $ in (\ref{eq:defEXP}). In other words, the notation $\mathbb{E}_{\vmesP} \left[ X \right]$ is another (shorter) notation for the integral $\int X \dif{ \vmesP}$.

\section{Stochastic Control Based on Free Energy and Relative Entropy Dualities}
\label{sec_dualities}

Let $( \mathrm{\Omega}, \vcalF) $ be a measurable space where $\mathrm{\Omega}$ is a non-empty set and $\vcalF \subseteq 2^{\mathrm{\Omega}}$ is a $\sigma$-algebra of subsets of $\mathrm{\Omega}$,
and let $\textbf{P}(\mathrm{\Omega})$ be the set of all probability measures defined on $( \mathrm{\Omega}, \vcalF) $.

\begin{defn}\rm
Let  $\vmesP \in \textbf{P}(\mathrm{\Omega})$ be a probability measure, $\vx=\vx(\omega),~\omega \in \mathrm{\Omega}$ be a random variable, $t, \rho \in \mathbb{R}$ be real numbers, and let $\vcalJ(\vx, t)$ be a measurable function. The \emph{Helmholtz free energy} of $\vcalJ(\vx, t)$ with respect to $\vmesP$ is defined by
\begin{equation}
\vcalE_{\vmesP} \left( \vcalJ ( \vx, t ); \rho  \right)  = \log \left( \int \exp\left( \rho \vcalJ( \vx, t) \right) \dif{\vmesP} \right) = \log\mathbb{E}_{\vmesP} \left[ \exp \left( \rho \vcalJ (\vx, t) \right) \right].
\end{equation}
\end{defn}

\begin{defn}\rm
Let $\vmesP, \vmesQ \in  \textbf{P}(\mathrm{\Omega})$ be two probability measures. The \emph{relative entropy} of $\vmesP$ with respect to $\vmesQ$ is defined as\footnote{Given two probability measures  $\vmesP$ and $\vmesQ$, we say that   $\vmesQ$ is\textit{ absolute continuous} with  $\vmesP$ and write   $\vmesQ \ll \vmesP $ if   $\vmesQ = 0 \Rightarrow \vmesP = 0 $,  see page 161 of \cite{Oksendal2003}.}:
\begin{equation} \mathbb{KL} \left( \vmesQ \|  \vmesP  \right) = \left\{
\begin{array}{l l}
  {\displaystyle \int\log \left(\frac{\dif{\vmesQ} }{\dif{\vmesP} } \right) \dif{\vmesQ}  }
 & \quad \mbox{if $\vmesQ \ll \vmesP $  and $ {\displaystyle \log\left( \frac{\dif{\vmesQ}}{ \dif{\vmesP}} \right) } \in L^{1}$,  }  \\
  +\infty & \quad \mbox{otherwise.}\\ \end{array}  \right.
  \end{equation}
\end{defn}

We will also consider the function $\xi(\vx, t)$, defined by
\begin{equation}
  \xi(\vx, t) = \onerho \vcalE_{\vmesP} \left( \vcalJ(\vx,t); \rho  \right)  =   \onerho \log \mathbb{E}_{\vmesP}\left[ \exp{\left( \rho \vcalJ(\vx,t) \right)}   \right].
\end{equation}
To derive the basic relationship between free energy and relative entropy \cite{Dai-Pra:1996fk}, we express the expectation $\mathbb{E}_{\vmesP}$ taken under the probability measure $ \vmesP$ as a function of the expectation  $ \mathbb{E}_{\vmesQ}  $ taken under the probability measure $ \vmesQ$. More precisely, we have:
\begin{equation*}
\mathbb{E}_{\vmesP} \left[ \exp\left( \rho \vcalJ(\vx, t) \right) \right] = \int  \exp{\left( \rho \vcalJ(\vx, t) \right)}  \frac{ \dif{\vmesP}}{ \dif{\vmesQ}}  \, \dif{\vmesQ}.
\end{equation*}

By taking the logarithm of both sides of the previous equation and by making use of Jensen's inequality \cite{Dai-Pra:1996fk}, it can be shown that:
\begin{equation}\label{JensenInequatily}
          \log \mathbb{E}_{\vmesP} \left[ \exp{\left( \rho \vcalJ(\vx, t) \right)}   \right]  \geq   \int  \rho { \vcalJ}(\vx,t) \, \dif{ \vmesQ}   - \mathbb{KL}\left( \vmesQ \| \vmesP  \right).
\end{equation}

Let $ \rho < 0 $. By multiplying both sides of (\ref{JensenInequatily}) with $ -1 / |\rho|  $, one obtains:
\begin{equation}\label{LowerBound}\boxed{
\xi(\vx, t) = - \frac{1}{| \rho| }   \vcalE_{\vmesP} \left( \vcalJ(\vx, t); \rho \right) \leq \mathbb{E}_{\vmesQ} \left[ \vcalJ(\vx,t) \right] + \frac{1}{|\rho|}  \mathbb{KL}\left( \vmesQ \| \vmesP \right)}
\end{equation}
where $\displaystyle \mathbb{E}_{\vmesQ} \left[ \vcalJ(\vx,t)  \right] =  \int{\cal{J}}(\vx,t) \, \dif{\vmesQ }$. The inequality (\ref{LowerBound}) provides us with a duality relationship between relative  entropy and  free energy. Essentially, one could define the following minimization problem:
\begin{equation} \label{Dual1}
-\frac{1}{|\rho|} \vcalE_{\vmesP} \left( \vcalJ(\vx,t); \rho  \right) = \inf_{\vmesQ \in \textbf{P}(\mathrm{\Omega})} \left( \mathbb{E}_{\vmesQ} \left[ \vcalJ(\vx,t) \right] + \frac{1}{|\rho|} \mathbb{KL} (\vmesQ \|\vmesP) \right).
\end{equation}

It can be shown that the infimum in (\ref{Dual1}) is attained at  $\vmesQ^{*}$, where
\begin{equation}\label{OptimalDistribution1}
\dif{\vmesQ}^{*}  = \frac{\exp{\left(- | \rho|  \vcalJ (\vx,t) \right) } }{ \int \exp{\left(- | \rho|  \vcalJ (\vx,t) \right)} \, \dif{\vmesP}} \, \dif{\vmesP}.
\end{equation}

A rather intuitive way of writing (\ref{LowerBound}) is to express it in the following form:
\begin{align}\label{FreeEnergyDefinitionInequality}
\underbrace{ \small{ -|\rho|^{-1} \vcalE_{\vmesP} \left( \vcalJ(\vx,t); \rho  \right)}}_\text{ \bf{Helmholtz~Free~Energy}} \leq \underbrace{ \text{State Cost} + |\rho|^{-1} \text{Information~Cost}}_\text{\bf{Non-Equilibrium~Free~Energy}}
\end{align}
where ``State Cost''  and ``Information Cost'' are defined as $ \mathbb{E}_{\vmesQ} \left[ \vcalJ(\vx, t)  \right]$ and
$\mathbb{KL}\left( \vmesQ \|\vmesP \right)$, respectively.

 In the next sections, we derive the form of  (\ref{Dual1}) for the case when $\vx$ is the state of a nonlinear stochastic differential equation affine in noise and control.
%
%
%

\subsection{Application of the Legendre Transformation to Stochastic Differential Equations }\label{sec:Nonlinear_AffineNoise}
We consider the general uncontrolled and controlled stochastic dynamics affine in noise as follows:
\begin{align}
	\dif{\vx} &=  \vA(\vx) \, \dif{t}   +   \vC(\vx)   \, \dif{ \vw }^{(0)},    \label{uncontrolled1} \\
    \dif{\vx} &=  \vF(\vx,\vu) \, \dif{t}   +   \vC(\vx)  \, \dif{\vw}^{(1)},     \label{controlled1}
\end{align}
where $ \vx \in \Real^{n} $ denotes the state of the system, $ \vu \in \Real^{m} $ denotes the control input, $ \vC(\vx) \in \Real^{n \times m} $ is the  diffusion matrix, $ \vF(\vx,\vu) \in \Real^{n} $ is the  drift  dynamics, and $ \vw^{(0),(1)} \in \Real^{m} $ are Wiener processes (Brownian motion). The upper-scripts $(0)$ and $(1)$ are used to distinguish  the two noise processes in the uncontrolled and controlled dynamics, respectively.  The drift term $\vA(\vx) \in \Real^{n} $  is defined by $ \vA(\vx) =  \vF(\vx,0)$. The diffusion matrix may be partitioned as $\vC (\vx) = \begin{bmatrix} \mathbf{0} & \vC_{c}^{\intercal}(\vx) \end{bmatrix}^{\intercal} $  where $\mathbf{0} \in \Real^{(n-m) \times m}$ and $\vC_{c} (\vx) \in \Real^{m \times m}$ is invertible. Similarly, the drift term in the controlled dynamics may be partitioned as $\vF (\vx,\vu) = \begin{bmatrix} \vF_{1}^{\intercal}(\vx,\vu) & \vF_{2}^{\intercal}(\vx,\vu) \end{bmatrix}^{\intercal} $ where $\vF_{1}(\vx,\vu) \in \Real^{m \times (n-m)}$ and $\vF_{2}(\vx,\vu) \in \Real^{m \times m}$; and the drift term in the uncontrolled dynamics may be partitioned as $\vA (\vx) = \begin{bmatrix} \vA_{1}^{\intercal}(\vx) & \vA_{2}^{\intercal}(\vx) \end{bmatrix}^{\intercal} $ where $\vA_{1}(\vx) \in \Real^{m \times (n-m)}$ and $\vA_{2}(\vx) \in \Real^{m \times m}$. The class of systems whose matrices can be partitioned as such contains rigid body, and multi body dynamics as well as kinematic models such as the ones considered in this work. Henceforth, for simplicity, we will assume that $m=n$. The case when $m < n$ can be treated similarly; see for instance \cite{Kushner1991}. Let $\VSigma (\vx) =  \vC(\vx)  \vC^{\intercal}(\vx) \in \Real^{m \times m}$ and
also define the following quantity:
\begin{equation*}
	\delta \vF (\vx, \vu)  = \vF (\vx,\vu) -   \vA (\vx)  = \vF (\vx,\vu) - \vF (\vx,0), \quad \forall \vx, \vu.
\end{equation*}
To the system (\ref{controlled1}) we also associated the state cost
\begin{equation} \label{eq:cost}
\vcalJ(\xx,t)  =  \mathrm{\Phi}(\vx (t_{\mathrm{f}})) +  \int_{t}^{t_{\mathrm{f}}} q(\vx (\tau), \tau) \, \dif{\tau}.
\end{equation}
With a slight abuse of notation we will also use $\vcalJ(\vx,t)$ to denote the value of $\vcalJ(\xx,t)$ along the trajectory $\xx$ starting from $\vx = \vx(t)$ at time $t$.
Expectations evaluated on trajectories generated by the uncontrolled dynamics and controlled dynamics will be represented by $\mathbb{E}_{\vmesP}[\,\cdot\,]$  and  $\mathbb{E}_{\vmesQ}[\,\cdot\,]$, respectively.
The following fact can be found in \cite{Kushner1991}.

\begin{proposition}
\rm
Given the measures $\vmesP, \vmesQ$ induced by the trajectories of (\ref{uncontrolled1}) and (\ref{controlled1}), respectively, the
\textit{Radon-Nikodym derivative} of $\vmesQ$ with respect to $\vmesP$  is defined by
\begin{align}
\frac{\dif{\vmesQ} }{\dif {\vmesP} } =& \, \exp \left(   \int_{t}^{t_{\mathrm{f}}} \delta \vF^{\intercal} (\vx(\tau),\vu(\tau))  \vC^{-1}(\vx(\tau)) \, \dif{\vw}^{(1)} (\tau) \right) + \nonumber \\
&\, \exp \left(  \int_{t}^{t_{\mathrm{f}}} \half  \delta \vF^{\intercal} (\vx(\tau),\vu(\tau))   \VSigma^{-1} (\vx(\tau))\, \delta \vF (\vx(\tau),\vu(\tau)) \,  \dif{\tau} \right).  \label{RadonNikodym}
\end{align}

\end{proposition}

Given equation (\ref{RadonNikodym}), the relative entropy term in (\ref{LowerBound}) takes the form:
\begin{align*}
 \frac{1}{|\rho|} \mathbb{KL} (\vmesQ \| \vmesP)  =& \, \mathbb{E}_{\vmesQ} \left[  \frac{1}{|\rho|} \int_{t}^{t_{\mathrm{f}}}   \delta \vF^{\intercal} (\vx(\tau),\vu(\tau)) \vC^{-1}(\vx(\tau)) \, \dif{\vw}^{(1) }  (\tau) \right] + \\
 &\, \mathbb{E}_{\vmesQ} \left[  \frac{1}{|\rho|} \int_{t}^{t_{\mathrm{f}}} \half  \delta \vF^{\intercal}(\vx (\tau),\vu(\tau)) \VSigma^{-1}(\vx(\tau)) \delta \vF(\vx(\tau),\vu(\tau))  \, \dif{\tau} \right]\\
 =& \, \mathbb{E}_{\vmesQ}\left[  \frac{1}{2|\rho|} \int_{t}^{t_{\mathrm{f}}} \delta \vF^{\intercal} (\vx (\tau),\vu(\tau))  \VSigma^{-1} (\vx(\tau))   \delta \vF(\vx(\tau),\vu(\tau)) \,\dif{\tau} \right],
\end{align*}
where the first term in the previous expression vanishes since  the expectations  term
$\small{\mathbb{E}_{\vmesQ} \left[  \delta \vF^{\intercal} (\vx(\tau),\vu(\tau)) \vC^{-1}(\vx(\tau)) \dif{\vw}^{(1) } (\tau) \right] }  $  becomes
\begin{equation}
  \mathbb{E}_{\vmesQ} \left[  \delta \vF^{\intercal} (\vx(\tau),\vu(\tau)) \vC^{-1}(\vx(\tau)) \right] \mathbb{E}_{\vmesQ} \left[ \dif{\vw}^{(1) } (\tau) \right] = 0, \quad  \forall \tau, t \le \tau \le t_{\mathrm{f}} 
\end{equation}

Substituting the previous expression of the Kullback-Leibler divergence into  (\ref{LowerBound}) one obtains
\begin{align*}
   -\frac{1}{| \rho| }
   \vcalE_{\vmesP} \left( \vcalJ(\vx, t); \rho \right) &\leq  \,  \mathbb{E}_{\vmesQ} \left[ \vcalJ (\vx, t)\right]  + \\
   &\hspace*{-5mm} \mathbb{E}_{\vmesQ} \left[ \frac{1}{2|\rho|}   \int_{t}^{t_{\mathrm{f}}} \delta \vF^{\intercal}(\vx(\tau),\vu(\tau))  \VSigma^{-1} (\vx(\tau))   \delta \vF(\vx(\tau),\vu(\tau)) \,  \dif{\tau} \right].
\end{align*}
The previous equation can be written in the form (\ref{FreeEnergyDefinitionInequality}) with state cost term defined as
\begin{equation}
	\mathbb{E}_{\vmesQ} \left[ \vcalJ(\vx,t)  \right], \\
\end{equation}
and information cost defined as
\begin{equation}
   \mathbb{E}_{\vmesQ} \left[ \frac{1}{2|\rho|}   \int_{t}^{t_{\mathrm{f}}}  \delta \vF^{\intercal}(\vx(\tau),\vu(\tau)) \VSigma^{-1} (\vx(\tau))\delta \vF(\vx(\tau),\vu(\tau)) \, \dif{\tau} \right].
\end{equation}

 %
%
%

Next, we further specialize the class of systems where (\ref{FreeEnergyDefinitionInequality}) is applied to, and discuss its connections to stochastic optimal control as in \cite{TheodorouYale2013, TheodorouCDC2013, Dai-Pra:1996fk}. To this end, let us  consider the special case of (\ref{uncontrolled1}) and (\ref{controlled1}) with uncontrolled  and controlled  stochastic dynamics of the following form, respectively:
\begin{equation}\label{uncontrolled}
	\dif{\vx} =  \vf(\vx) \, \dif{t} +\frac{1}{\sqrt{|\rho|}}  \vB(\vx)  \, \dif{\vw}^{(0)},
\end{equation}
\begin{equation}\label{controlled}
	\dif{\vx} =  \vf(\vx) \, \dif{t}   + \vB(\vx) \left( \vu \, \dif{t} + \frac{1}{ \sqrt{|\rho|}} \, \dif{\vw}^{(1)} \right),
\end{equation}
where $ \vx \in \Real^{n} $ denotes the state of the system, $ \vB(\vx) \in \Real^{n \times m} $  is the control/diffusion matrix, $ \vf(\vx) \in \Real^{n} $ is the passive dynamics,  $\vu \in \Real^{m}  $ is the control vector and $ \vw^{(0),(1)}$  are $m$-dimensional Wiener noise processes. 

For the dynamics in  (\ref{uncontrolled}) and (\ref{controlled}) the  form of the Radon-Nikodym derivative in (\ref{RadonNikodym}) can be computed as follows. Noticing  that
$ \delta \vF(\vx,\vu)  =   \vB(\vx) \vu $,  $\vC(\vx) =  \vB(\vx) /\sqrt{|\rho|}   $ and $ \VSigma (\vx)= \vB(\vx) \vB^{\intercal}(\vx) / |\rho|$, and substituting these expressions in  (\ref{RadonNikodym}) yields
\begin{equation} \label{Randon}
	\frac{\dif{\vmesQ}}{\dif{\vmesP}} = \exp\left( |\rho| \eta(\vu,t)  \right)  \quad \text{and} \quad  \frac{\dif{\vmesP}}{\dif{\vmesQ}}  =  \exp\left( -|\rho| \eta(\vu,t)  \right),
\end{equation}
where  $\eta(\vu,t)$ is given by:

\begin{equation} \label{Girsanov}
	\eta(\vu,t) = \frac{1}{2} \int_{t}^{t_{\mathrm{f}}}   \vu^{\intercal}(\tau)  \vu(\tau) \, \dif{\tau} + \frac{1}{\sqrt{|\rho|}} \int_{t}^{t_{\mathrm{f}}}  \vu^{\intercal}(\tau)  \, \dif{\vw}^{(1)}.
\end{equation}

%
Substitution of (\ref{Randon}) and (\ref{Girsanov}) into  inequality (\ref{LowerBound}) yields the following result:
\begin{equation} \label{LowerBound0}
-  \frac{1}{|\rho|} \log \mathbb{E}_{\vmesP} \left[ \exp \left(- |\rho| \vcalJ(\vx,t) \right)   \right] \leq  \mathbb{E}_{\vmesQ} \left[  \vcalJ (\vx,t) + \frac{1}{|\rho|} \eta(\vu,t)  \right].
\end{equation}
The expectation on the right side of the inequality in (\ref{LowerBound0}) is further simplified as follows:
\begin{equation}\label{LowerBound1}
 \underbrace{   -  \frac{1}{|\rho|}  \log \mathbb{E}_{\vmesP} \left[ \exp{\left( - |\rho| \vcalJ(\vx,t) \right)}   \right]}_{\xi(\vx,t)}  \leq  \underbrace{\mathbb{E}_{\vmesQ}  \left[  \vcalJ (\vx,t)  +   \half \int_{t}^{t_{\mathrm{f}}}  \vu(\tau)^{\intercal}  \vu(\tau)  \, \dif{\tau}  \right].}_{\text{Total Cost}}
\end{equation}

The right-hand side term in the above inequality   corresponds to the cost function of a  stochastic optimal control problem that is bounded from below by the free energy. Surprisingly, inequality (\ref{LowerBound1}) was derived  without  relying on any principle of optimality.  Inequality  (\ref{LowerBound1}) essentially defines a  minimization process in which the right-hand side part  of the inequality is  minimized with respect  to $ \eta(\vu,t) $ and therefore with respect to the corresponding control $\vu$. At the minimum, when $ \vu = \vu^{*} $, the right-hand side of inequality  in (\ref{LowerBound1}) attains its optimal value $ \xi(\vx,t) $. Under the optimal control $ \vu^{*}  $,  and according to   (\ref{OptimalDistribution1}), the corresponding optimal distribution takes the form

\begin{equation}\label{OptimalDistribution2}
 \dif{\vmesQ}^{*}= \frac{\displaystyle \exp{\Big( - |\rho|  \mathrm{\Phi}(\vx (t_{\mathrm{f}}))    \Big)} \exp{\left(- |\rho| \int_{t}^{t_{\mathrm{f}}}  q(\vx (\tau), \tau) \, \dif{\tau}  \right) }  }{ \displaystyle \int \exp{\Big( - |\rho|  \mathrm{\Phi}(\vx (t_{\mathrm{f}}))    \Big)} \exp{\left( -|\rho| \int_{t}^{t_{\mathrm{f}}}  q(\vx (\tau), \tau) \, \dif{\tau}  \right) \dif{\vmesP} }} \, \dif{\vmesP}.
\end{equation}


The work \cite{TheodorouYale2013, TheodorouCDC2013} inspired by early mathematical developments  in control theory  \cite{Dai-Pra:1996fk, Fleming:1995},   has shown   that the  value function   $\xi(\vx,t)$ in (\ref{LowerBound1}) satisfies the Hamilton-Jacobi-Bellman equation and it has made the connection with more recent work in machine learning \cite{Kappen2012, todorov2009efficient} on Kullback-Leibler and path integral control.

\subsection{Connection with Dynamic Programming (DP)}

An important question that arises is: What is the link  between \eqref{LowerBound1} and the principle of optimality in dynamic programming?
To address this question,  we show that $ \xi(\vx,t) $ satisfies the Hamilton-Jacobi-Bellman (HJB) equation associated with the optimal control problem
(\ref{controlled})-(\ref{eq:cost}) and  hence, $\xi(\vx,t) $ is the corresponding value function of the following minimization problem
\begin{align} \label{opt_control_problem}
\xi (\vx,t) =& \min_{ \substack{\vu(\tau) \\ t \leq \tau \leq t_{\mathrm{f}}} } \mathbb{E}_{\vmesQ} \left[ \mathrm{\Phi}(\vx(t_{\mathrm{f}})) + \int_{t}^{t_{\mathrm{f}}} \left( q(\vx(\tau), \tau) + \half \vu^{\intercal} (\tau) \vu(\tau) \right) \dif{\tau}  \right] \nonumber\\
=& \min_{ \substack{\vu(\tau) \\ t \leq \tau \leq t_{\mathrm{f}}} } \mathbb{E}_{\vmesQ} \left[ \vcalJ(\vx, \tau) + \half \int_{t}^{t_{\mathrm{f}}}  \vu^{\intercal} (\tau) \vu(\tau)  \, \dif{\tau}  \right],
\end{align}
where the expectation is computed over the trajectories of (\ref{controlled}). To see this, we introduce $\mathrm{\Psi}(\vx,t) \triangleq \mathbb{E}_{\vmesP} \left[ \exp \left( \rho \vcalJ(\vx,t) \right) \right]$ and apply the Feynman-Kac lemma \cite{Friedman75}
to arrive at the backward Chapman-Kolmogorov partial differential equation (PDE)
\begin{align}
- \partial_{t}\mathrm{\Psi}(\vx,t) =& \, - |\rho|  q(\vx,t) \mathrm{\Psi}(\vx,t)+  \vf^{\intercal}(\vx) \nabla \mathrm{\Psi}_{\vx}(\vx,t)  \nonumber\\
& \qquad \qquad + \frac{1}{2 |\rho|}\Tr \left(\nabla \mathrm{\Psi}_{\vx \vx}(\vx,t)  \vB(\vx) \vB(\vx)^{\intercal}   \right) \label{KolmogorovPDE0}
\end{align}
with boundary condition $\mathrm{\Psi}(\vx(\tf),\tf) = \exp{\big(- |\rho| \mathrm{\Phi}(\vx(\tf)\big)}$,
which governs the evolution of $\mathrm{\Psi}(\vx,t) $ along the trajectories of (\ref{controlled}) subject to $\vx = \vx(t)$. 
Since $ \xi(\vx, t) = - \log \mathrm{\Psi}(\vx,t) / | \rho|$,
it follows that
\begin{align*}
\partial_{t} \mathrm{\Psi}(\vx,t) &=-| \rho| \mathrm{\Psi}(\vx,t) \partial_{t} \xi(\vx,t), \\
\nabla \mathrm{\Psi}_{\vx}(\vx,t) &= - \vert \rho \vert \mathrm{\Psi}(\vx,t)  \nabla \xi_{\vx} (\vx,t), \\
\nabla \mathrm{\Psi}_{\vx \vx}(\vx,t) &= |\rho | \mathrm{\Psi}(\vx,t) \nabla \xi_{\vx \vx}(\vx,t) -  \vert \rho \vert ^2 \mathrm{\Psi} (\vx,t) \nabla\xi_{\vx}(\vx,t)  \nabla\xi_{\vx}^{\intercal}(\vx,t).
\end{align*}

\noindent In this case, it can be  shown that $ \xi(\vx,t)$ satisfies the nonlinear PDE
\begin{align}\label{HJBEquation2}
- \partial_{t} \xi(\vx,t) &=  q(\vx,t) +  \nabla\xi_{\vx}^{\intercal}(\vx,t) \vf(\vx)  - \half \nabla\xi_{\vx}^{\intercal}(\vx,t)\vB(\vx)  \vB^{\intercal} (\vx)  \nabla \xi_{\vx} (\vx,t) \nonumber \\
& \qquad\qquad + \frac{1}{2 |\rho|} \Tr \left( \nabla \xi_{\vx \vx}(\vx,t)  \vB(\vx)  \vB^{\intercal}(\vx)  \right),
\end{align}
subject to  the boundary condition $\xi(\vx(\tf),\tf) =  \mathrm{\Phi}(\vx(\tf))$.
The nonlinear PDE \eqref{HJBEquation2} corresponds to the HJB equation associated with the optimal control problem~(\ref{opt_control_problem}) and hence $\xi(\vx,t)$ is the corresponding minimizing value function~\cite{Stengel1994}. It is important to note, however, that the principle of optimality was not used to derive \eqref{HJBEquation2}.

\subsection{Path Integral Control with Initial Sampling Policies} \label{sec_iterative_case}

According to (\ref{LowerBound1}), in order to find the value function $ \xi(\vx,t) $, sampling of trajectories under the uncontrolled dynamics  is performed, and  the  left-hand side of (\ref{LowerBound1}) is evaluated on these trajectories. However, in high-dimensional spaces, it is desirable to steer sampling towards specific areas of the state space. To do so, we have to incorporate  an initial control policy into the uncontrolled dynamics. Therefore, instead of sampling from the uncontrolled dynamics~(\ref{uncontrolled}), we sample, instead, based on the stochastic  dynamics:
\begin{equation}  \label{DynamicsRRT}
   \dif{\vx} =  \vf(\vx) \, \dif{t}   + \vB(\vx) \left( \vu_{\mathrm{in}} \, \dif{t} + \frac{1}{ \sqrt{|\rho|}}  \, \dif{\vw}^{(1)} \right),
\end{equation}
where $\vu_{\mathrm{in}} $ is an initial control policy.   In \cite{TheodorouYale2013, TheodorouCDC2013}, the authors derived an  iterative PI control without relying on previous policy parameterizations. More  precisely, when sampling from the dynamics (\ref{DynamicsRRT}) the work in \cite{TheodorouCDC2013} and \cite{TheodorouYale2013} showed that the value function  $ \xi(\vx,t) $ is expressed as


\begin{align*}
	\xi(\vx,t) = - \frac{1}{| \rho |} \log\left( \int  \exp\left( - |\rho|  S \left(\vx,\vu_{\mathrm{in}}(\vx,t), t \right) \right)  \dif{\vmesQ_{\mathrm{in}} } \right)
\end{align*}
\noindent where the term $S(\vx,\vu_{\mathrm{in}})$ is defined as
\begin{align} \label{pathcost}
S(\vx,\vu_{\mathrm{in}}) =&\, \underbrace{ \mathrm{\Phi(\vx(t_{\mathrm{f}}))} + \int_{t}^{t_{\mathrm{f}}} q(\vx(\tau), \tau) \, \dif{\tau} }_{\vcalJ(\vx,t)}  +  \nonumber\\
&\, \underbrace{ \frac{1}{2} \int_{t}^{t_{\mathrm{f}}} \vu_{\mathrm{in}}^{\intercal}(\tau) \vu_{\mathrm{in}}(\tau) \,\dif{\tau} + \frac{1}{\sqrt{|\rho|} }  \int_{t}^{t_{\mathrm{f}}} \vu_{\text{in}}^{\intercal}(\tau)  \, \dif{\vw}^{(1)}(\tau) }_{\eta(\vu_{\mathrm{in}},t)},
\end{align}
where the term $ \eta(\vu_{\mathrm{in}},t)$ appears due to sampling based on  the dynamics (\ref{DynamicsRRT}), while the term  $  {  \vcalJ(\vx,t)} $ is the state-dependent part of the  total cost function in (\ref{LowerBound1}). The path integral control is now expressed as  \cite{TheodorouYale2013}
\begin{equation} \label{Update_Control1}
   \vu_{\mathrm{PI}}(\vx, t) \, \dif{t} =    \vu_{\mathrm{in}}(\vx, t) \, \dif{t} + \delta \vu(\vx, t),
\end{equation}
where the term $ \delta\vu(\vx,t)$ is defined by
\begin{equation}\label{OptimalCorrection}
\delta \vu(\vx,t)  = \frac{1}{\sqrt{|\rho|}} \mathbb{E}_{\vmesQ^{*}}\! \left[  \dif{\vw}^{(1)} \right] = \frac{1}{\sqrt{|\rho|}}  \int  \, \dif{\vw}^{(1)}  \, \dif{ \vmesQ}^{*},
\end{equation}
and where the expectation is taken under the optimal probability
\begin{equation}
      \dif{ \vmesQ}^{*} = \frac{ \exp \left( -|\rho|  S(\vx,\vu_{\mathrm{in}}) \right)  }{ \displaystyle \int  \exp \left(- |\rho|   S(\vx,\vu_{\mathrm{in}}) \right)  \, \dif{\vmesQ}_{\mathrm{in}}    }  \,  \dif{\vmesQ}_{\mathrm{in}} .
\end{equation}
During implementation, equation  (\ref{ApproximateOptimalCorrection}) is approximated as
\begin{equation}\label{ApproximateOptimalCorrection}
      \delta \vu (\vx,t) =  \frac{1}{\sqrt{|\rho|}}  \sum_{k=1}^{\mathrm{\# traj}} p_{k} \dif{\vw}^{(1)} (\omega_k) ~~ \text{with} ~~ p_{k} = \frac{\exp \left(- |\rho|  S(\vx_{k},\vu_{\mathrm{in}})\right) } { \sum_{\ell=1}^{\mathrm{\# traj}} \exp \left(- |\rho|  S(\vx_{\ell},\vu_{\mathrm{in}}) \right) }
\end{equation}
The  initial policy $\vu_{\mathrm{in}} $   can be  a suboptimal control law, a hand-tuned PD, PID control, or feedforward control. In this paper, we  consider a feedforward control given by  the \AlgRRT{} algorithm as the initial control policy. In this case, the \AlgRRT{}-based optimal path integral control takes the form
\begin{equation} \label{Update_Control1}
   \vu_{\mathrm{PI}}(\vx,t) \,\dif{t} =    \vu_{\mathrm{RRT}}(t) \,\dif{t} + \delta \vu(\vx,t).
\end{equation}
In the next section, we discuss how to use the \AlgRRT{} algorithm to compute the initial control policy $\vu_{\mathrm{RRT}}$.

\section{Trajectory Sampling via Sampling-based Algorithms}

As shown in the previous sections, sampling of useful trajectories from the unforced dynamics can be a tedious task. This issue can be addressed by first computing a ``good enough'' initial trajectory and then sampling local trajectories in the neighborhood of this trajectory. In the proposed approach, we use a probabilistic algorithm to compute an initial trajectory quickly. Probabilistic methods have proven to be very efficient for the solution of motion planning problems with dynamic constraints in high dimensional search spaces. Among them, Rapidly-exploring Random Trees (RRTs)~\cite{LaValle2001, LaValle2006, Choset2005} are among the most popular for solving single query motion planning problems. The main body of the \AlgRRT{} algorithm is given in Algorithm~\ref{alg:rrt}.

In the proposed approach, we leverage the speed and exploration capabilities of the \AlgRRT{} algorithm to compute an initial policy quickly by making a minor modification of the \AlgRRT{} primitive procedures. Since both final time and final state are given, the search space is formed by adding an additional time dimension $T$ to the state space $\scX$. Our search space, goal set and free space are thus  defined as $\scZ = \scX \times T$, $ \scZGoal = \scXGoal \times \cTGoal$, and $\Zfree = \scZ \setminus \scZGoal$, respectively. The \AlgRRT{} algorithm is then run to find a trajectory starting from an initial point $\lZInit = (\lXInit, \lTInit)$ to the goal set $\scZGoal$ while avoiding the obstacles in $\scX$. The primitive procedures used by the \AlgRRT{} algorithm are given below:

\emph{Sampling}: $\PrcSample : \naturals \to \Zfree$ returns independent, identically distributed (i.i.d) samples from $\Zfree$.

\emph{Nearest neighbor}: $\PrcNearest$  returns a point from a given finite set $V$, which is the point closest to a given point $\vz$ in terms of a given distance function.

\emph{Steering}: Given two points $\vz_{1}$ and $\vz_{2}$ in $\Zfree$, $\PrcSteer$ extends $\vz_{1}$ towards $\vz_{2}$ by sampling trajectories from the unforced dynamics of the system. Specifically, the procedure samples a set of trajectories emanating from $\vz_{1}$ and returns the closest end point of this set of trajectories with respect to a given distance function.

\emph{Collision checking}: Given a trajectory $\vsigma$, the Boolean function $\PrcObstacleFree(\sigma)$ checks whether $\vsigma$ belongs to $\Zfree$ or not. It returns $\True$ if the trajectory is a subset of $\Zfree$, i.e., $\vsigma \subset \Zfree$, and $\False$ otherwise.

\emph{Graph extension}: $\PrcExtend$ is a function that extends the nearest vertex of the graph $\graph{G}$ toward the randomly sampled point $\zrand$. Since time always flows in forward direction, we make sure that $\PrcExtend$ computes valid connections, i.e., it returns false if the time value of $\zrand$ is less than that of the nearest vertex in the graph. The $\PrcExtend$ procedure of the \AlgRRT{} algorithm is shown in Algorithm~\ref{alg:extend_rrt}.

\include{./algorithms/rrt.tex}

\include{./algorithms/extend_rrt.tex}

The body of the path-integral based \AlgRRT{} algorithm is shown in Algorithm~\ref{alg:rrt_pi}. It runs in a receding horizon fashion, that is, it computes a ``good enough'' control input and executes the first portion of the control signal at each time step. The algorithm starts by initializing the current time and state with the initial values in Lines 2-3. The algorithm then computes an initial policy in Line 5 by using the \AlgRRT{} algorithm. The steering procedure in the \AlgRRT{} algorithm is slightly modified in order to sample dynamically feasible trajectories. The steering procedure first samples a fixed number of trajectories from the unforced dynamics and then chooses the one that has the closest terminal state towards the desired point. Once a trajectory that reaches the goal set has been computed, the corresponding trajectory $\vsigma_{\AlgRRT}$, along with control the signal $\mathbf{u}_{\AlgRRT}$, are extracted from the computed data structure in Line 6. Then, the algorithm proceeds by locally sampling trajectories around $(\vsigma_{\AlgRRT}, \mathbf{u}_{\AlgRRT})$ and computes the variation in the control $\delta \vu(\vx,t) $ according to (\ref{OptimalCorrection}) by using information of local trajectories. Since we have $M$ number of local trajectories, the expecation in~(\ref{OptimalCorrection}) is numerically approximated by using the expression in~(\ref{ApproximateOptimalCorrection}). For each local trajectory $\vsigma_{k}$, a cost value is computed as $S(\vsigma_{k}, \vu_{\AlgRRT})$ and its desirability value is computed by exponentiating the corresponding cost value, i.e., $d_{k} = \exp(-|\rho| S(\vsigma_{k}, \vu_{\AlgRRT}))$. Then, the variation term in control $\delta \vu (\vx,t)$ is computed by taking the weighted average of all noise profiles which create the local trajectories and the weight of each trajectory is computed as the normalized desirability value, i.e., $p_{k} = d_{k}/\Sigma_{\ell = 1}^{N} d_{\ell}$. The iteration of the algorithm is completed by executing the first $\tau$ times of the computed control signal and the algorithm keeps repeating the same steps until the final time is reached.

\include{./algorithms/rrt_pi.tex}

\FloatBarrier

\section{Numerical Simulations}
	
In this section, we present a series of simulated experiments using a kinematic car model. We are interested in controlling a vehicle, whose motion is described by the following kinematic equations:
\begin{equation}~\label{eqn:kinematic_car_model}
\dot{x} = v \cos{\theta}, \quad \dot{y} = v \sin{\theta}, \quad \dot{\theta} = w / r
\end{equation}
where $x$, $y$ are the Cartesian coordinates of a reference point of the vehicle, $v$ is its speed, $w$ is the control input and $r$ is a positive constant. We assume that the admissible control inputs, are restricted by $w \in [-1,1]$. We would like to find an optimal policy for the heading rate $w$ to move the vehicle from a given initial configuration $(\xinitial, \yinitial, \thetainitial)^{\intercal}$ to a final configuration $(\xfinal, \yfinal, \thetafinal)^{\intercal}$ within some fixed final time $t_{\mathrm{f}}$.

Let $\vx_{1} = x$, $\vx_{2} = y$, $\vx_{3} = \theta$ be the states and $\vu = w$ be the control input of the system. Then (\ref{eqn:kinematic_car_model}) can be rewritten as
\begin{equation}\label{eqn:kinematic_car_model2}
\dot{\vx}_{1} = v \cos{\vx_{3}}, \quad \dot{\vx}_{2} = v \sin{\vx_{3}}, \quad \dot{\vx}_{3} = \vu / r.
\end{equation}
Assuming the system is subjected to noise of intensity $\alpha$ in the control channel, (\ref{eqn:kinematic_car_model2}) can be written in the standard form
\begin{equation}
\begin{pmatrix} \dif{\vx}_{1} \\ \dif{\vx}_{2} \\ \dif{\vx}_{3} \end{pmatrix} = \begin{pmatrix} v \cos \vx_{3} \\ v \sin \vx_{3} \\ 0 \end{pmatrix}  \dif{t} + \begin{pmatrix} 0 \\ 0 \\ \sfrac{1}{ r} \end{pmatrix} (\vu \, \dif{t} + \alpha \, \dif{\vw}),
\end{equation}
where $\vf$, $\vB$ and $\rho$ in (\ref{DynamicsRRT}) are defined as follows
\begin{equation*}
 \vf(\vx) = \begin{pmatrix} v \cos \vx_{3} \\ v \sin \vx_{3} \\ 0 \end{pmatrix}, \quad \vB = \begin{pmatrix} 0 \\ 0 \\  \sfrac{1}{r} \end{pmatrix}, \quad \rho = -\frac{1}{\alpha^{2}}.
\end{equation*}

The following parameters were used in the numerical simulations:
$\vx_{0} = \begin{pmatrix}-9 & 0 & 0 \end{pmatrix}^{\intercal}$, $t_{0} = 0$, $\vx_{\mathrm{f}} = \begin{pmatrix} 9 & 0 & 0 \end{pmatrix}^{\intercal}$, $t_{\mathrm{f}} = 10$, $\dif{t} = 0.1$, $v = 2.0$.

\subsection{Example 1: Single-slit Obstacle}

\begin{figure*}[htp]
\centering
	\mbox{  
    \subfigure[]{\scalebox{0.30}{\includegraphics[trim = 4.5cm 7.6cm 4.0cm 7.3cm, clip =
          true]{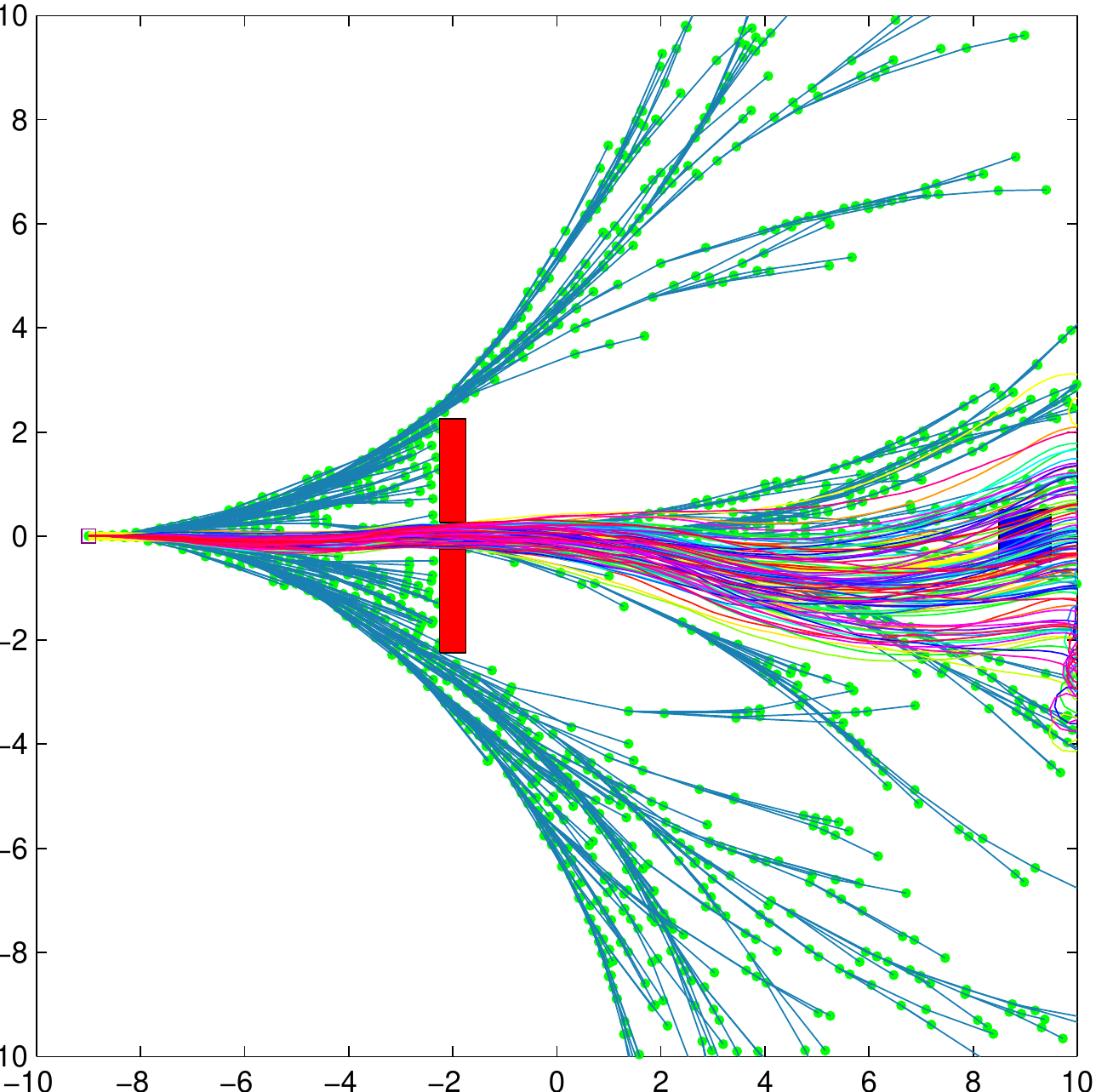}} \label{figure:pt1_alpha025_fr1}} 		
    \subfigure[]{\scalebox{0.30}{\includegraphics[trim = 4.5cm 7.6cm 4.0cm 7.3cm, clip =
          true]{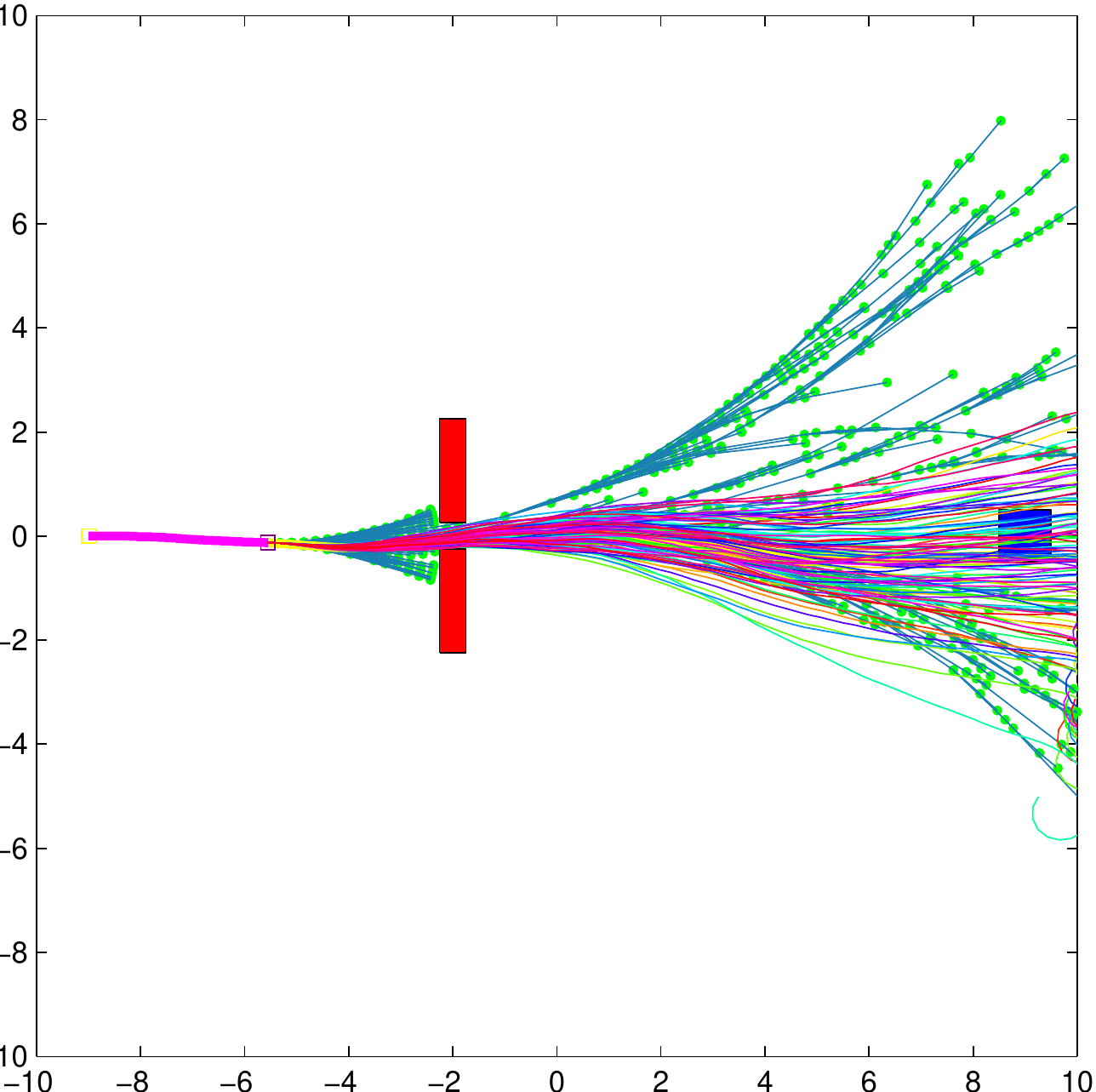}} \label{figure:pt1_alpha025_fr2}}
	\subfigure[]{\scalebox{0.30}{\includegraphics[trim = 4.5cm 7.6cm 4.0cm 7.3cm, clip =
          true]{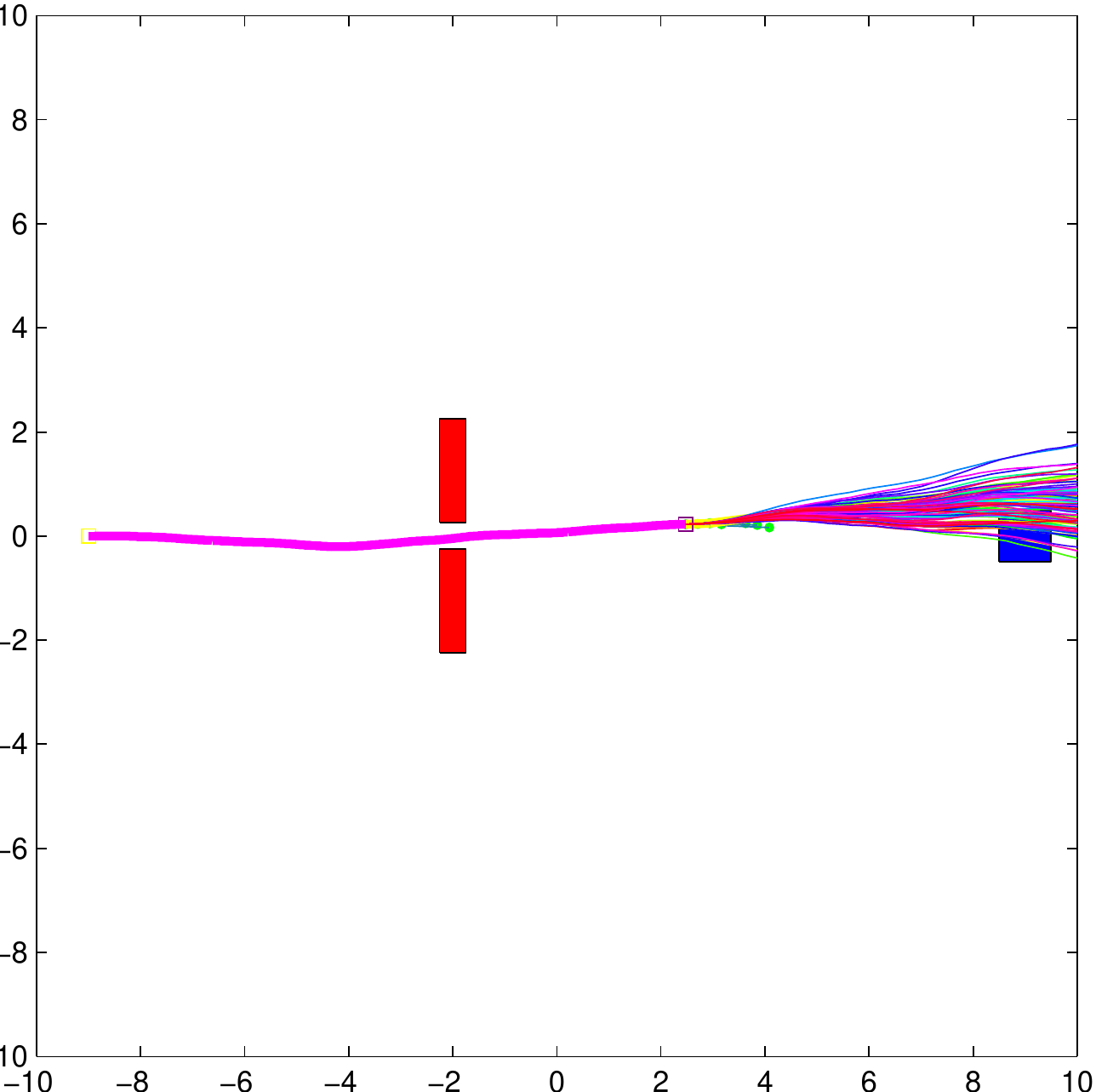}} \label{figure:pt1_alpha025_fr3}}
          }
    \mbox{  
    \subfigure[]{\scalebox{0.30}{\includegraphics[trim = 4.5cm 7.6cm 4.0cm 7.3cm, clip =
          true]{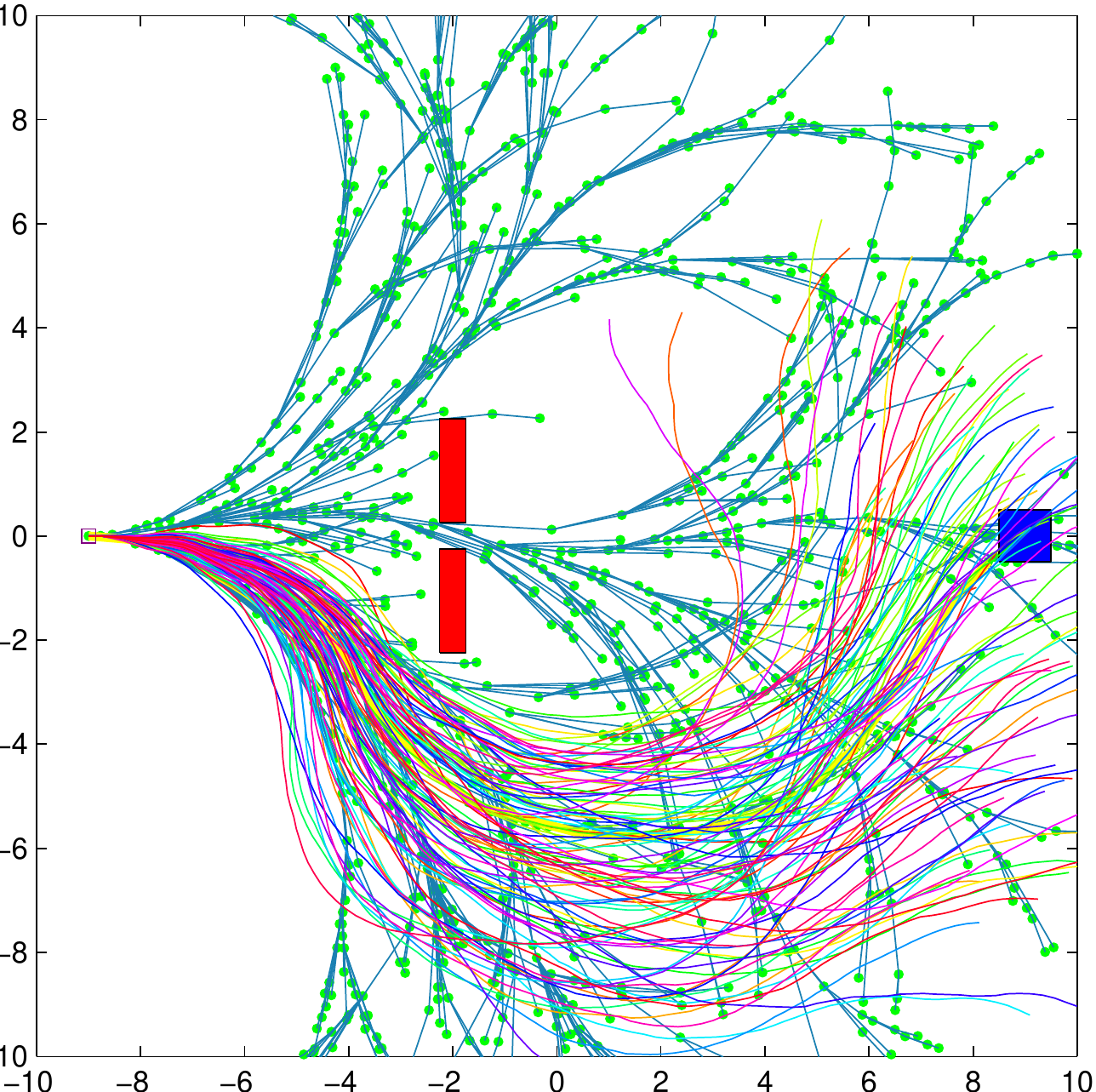}} \label{figure:pt1_alpha050_fr1}} 		
    \subfigure[]{\scalebox{0.30}{\includegraphics[trim = 4.5cm 7.6cm 4.0cm 7.3cm, clip =
          true]{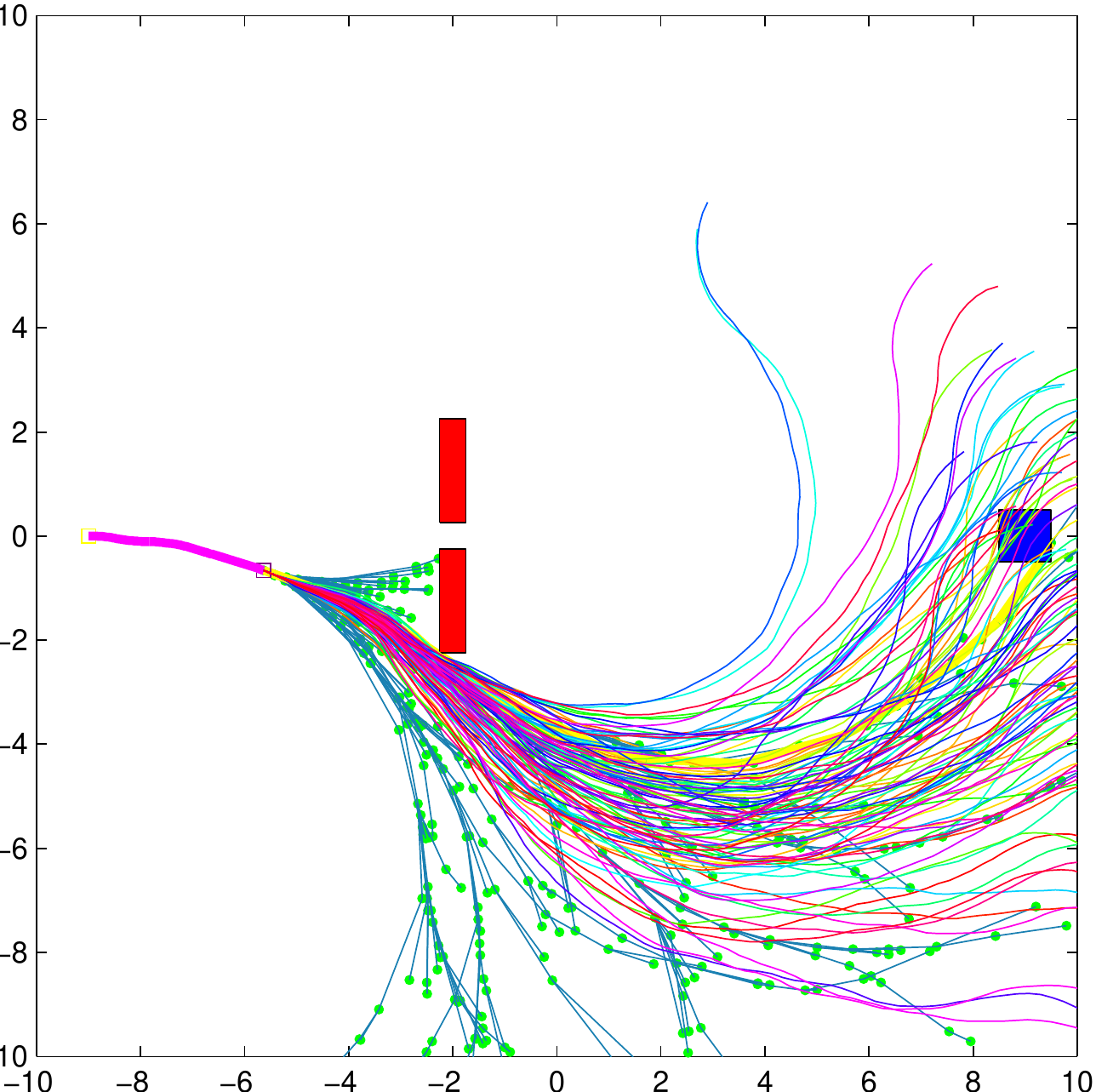}} \label{figure:pt1_alpha050_fr2}}
	\subfigure[]{\scalebox{0.30}{\includegraphics[trim = 4.5cm 7.6cm 4.0cm 7.3cm, clip =
          true]{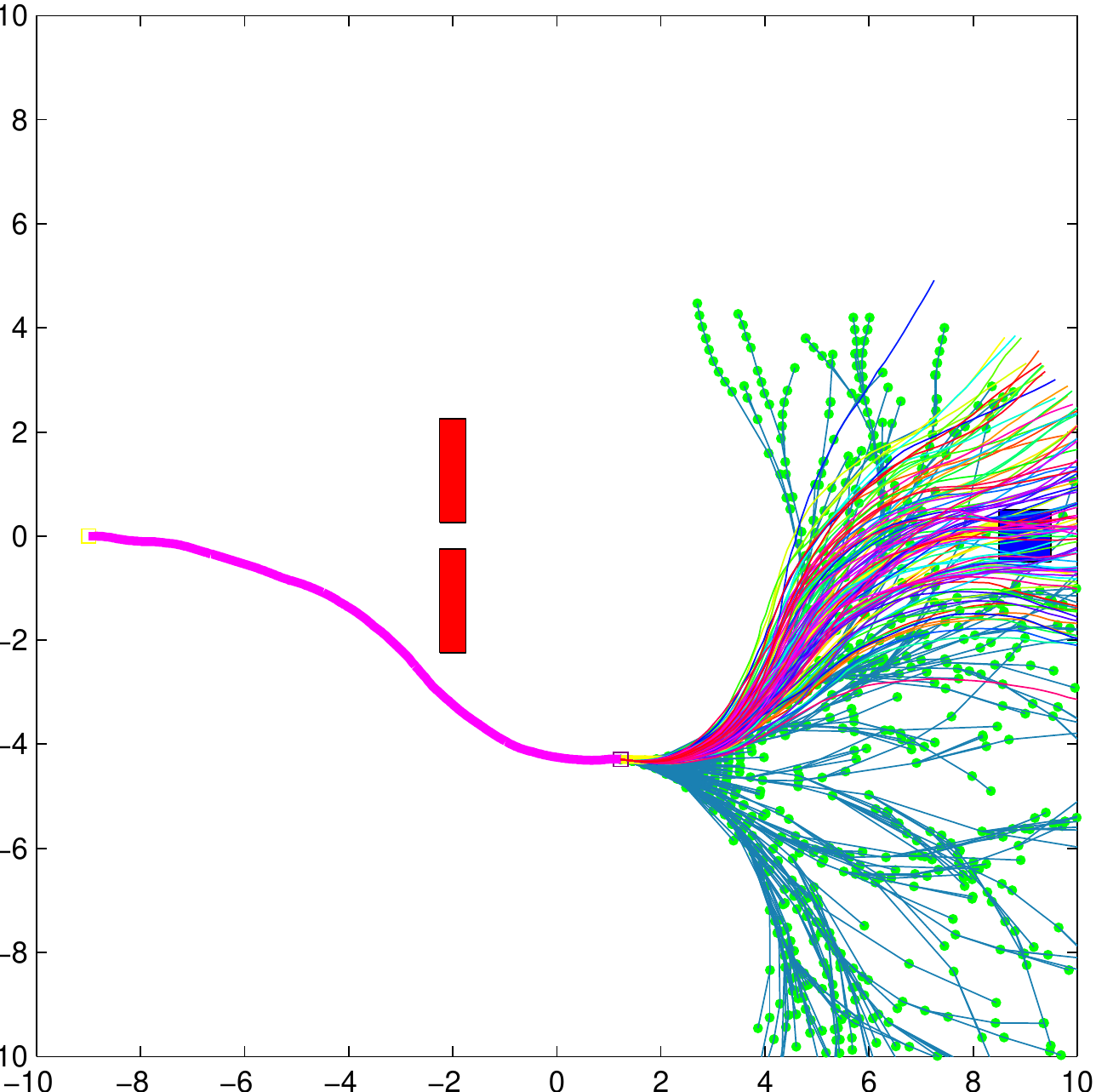}}\label{figure:pt1_alpha050_fr3}}
          }
	\mbox{  
    \subfigure[]{\scalebox{0.30}{\includegraphics[trim = 4.5cm 7.6cm 4.0cm 7.3cm, clip =
          true]{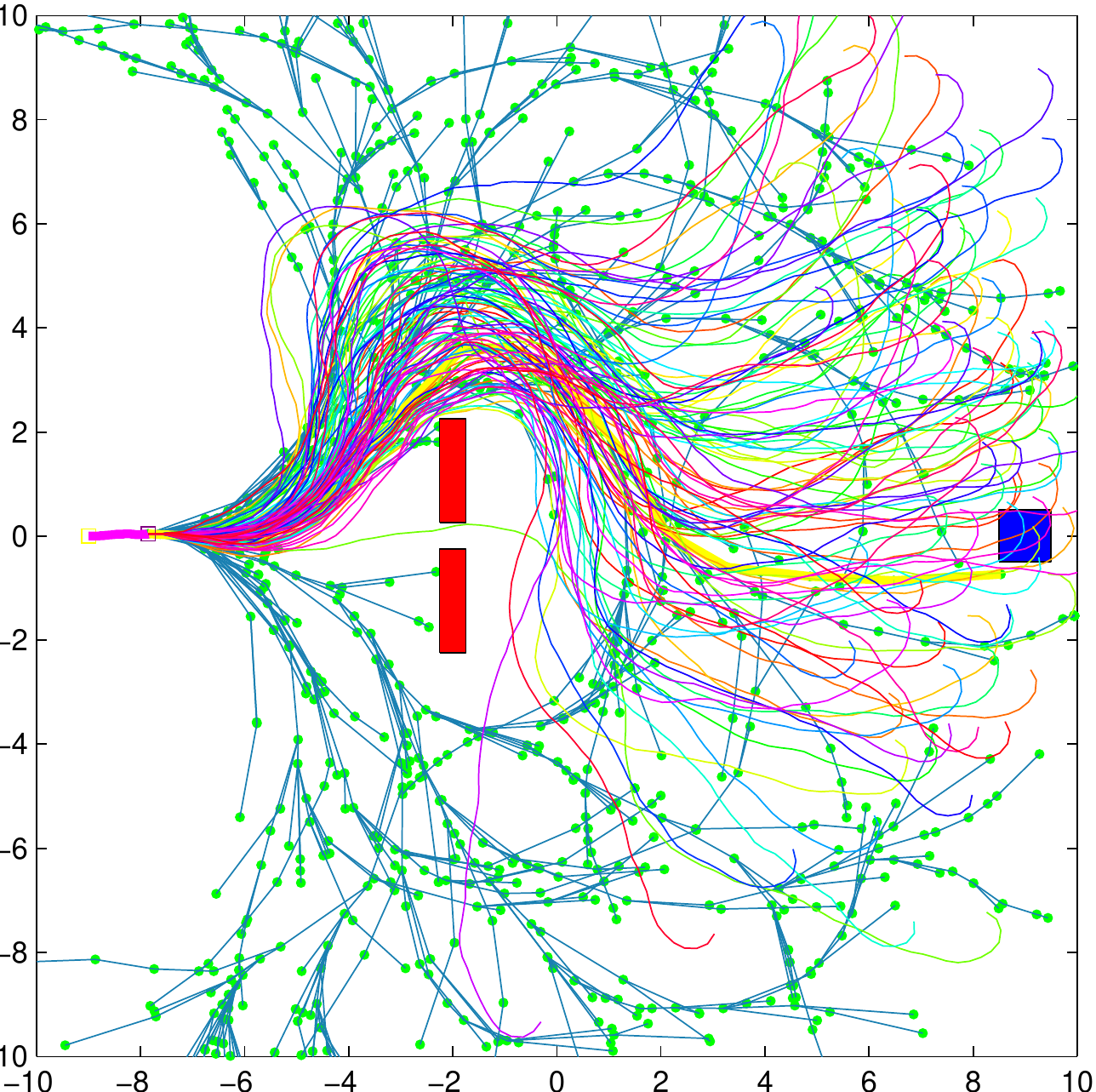}} \label{figure:pt1_alpha100_fr1}} 		
    \subfigure[]{\scalebox{0.30}{\includegraphics[trim = 4.5cm 7.6cm 4.0cm 7.3cm, clip =
          true]{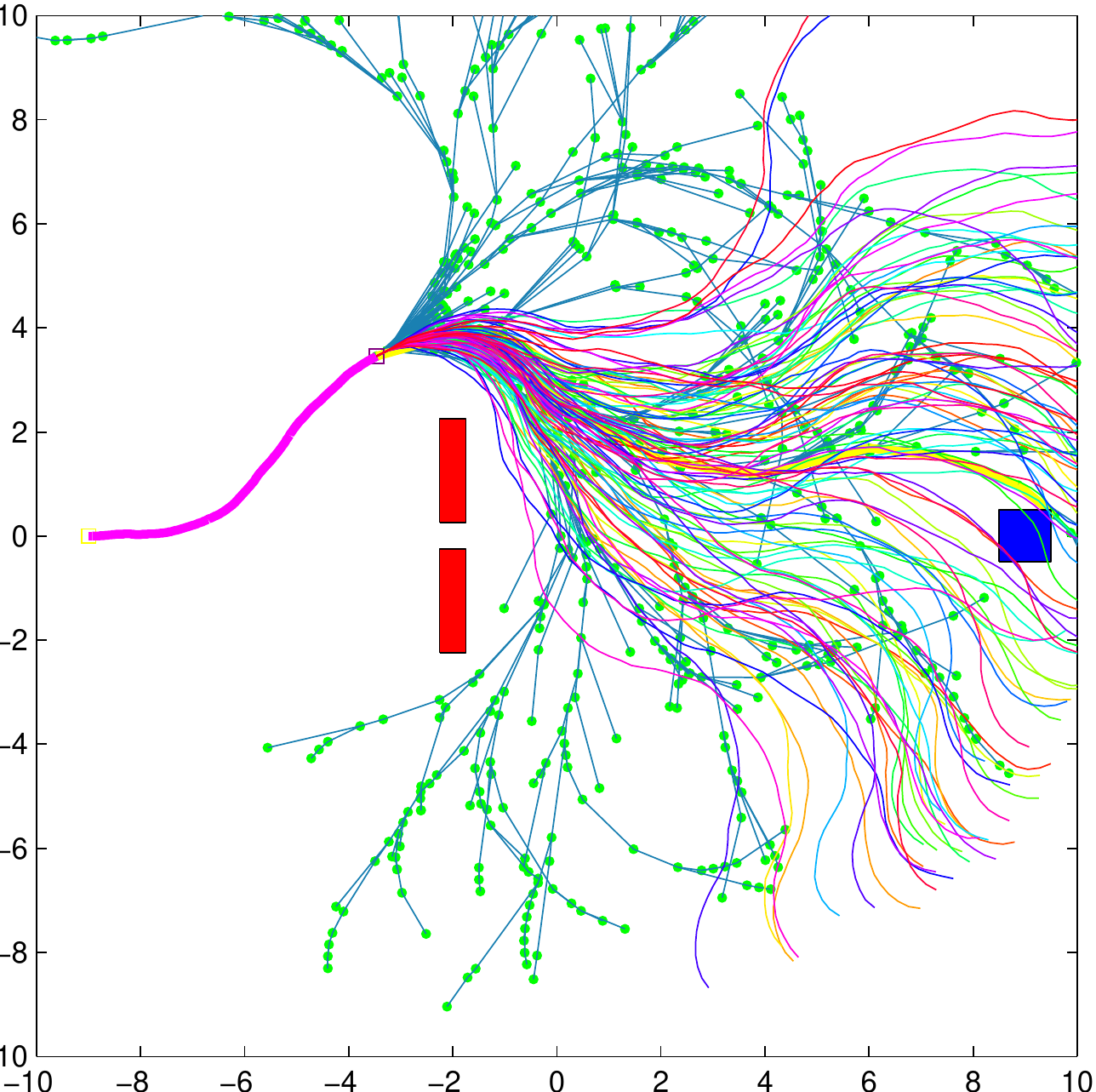}} \label{figure:pt1_alpha100_fr2}}
	\subfigure[]{\scalebox{0.30}{\includegraphics[trim = 4.5cm 7.6cm 4.0cm 7.3cm, clip =
          true]{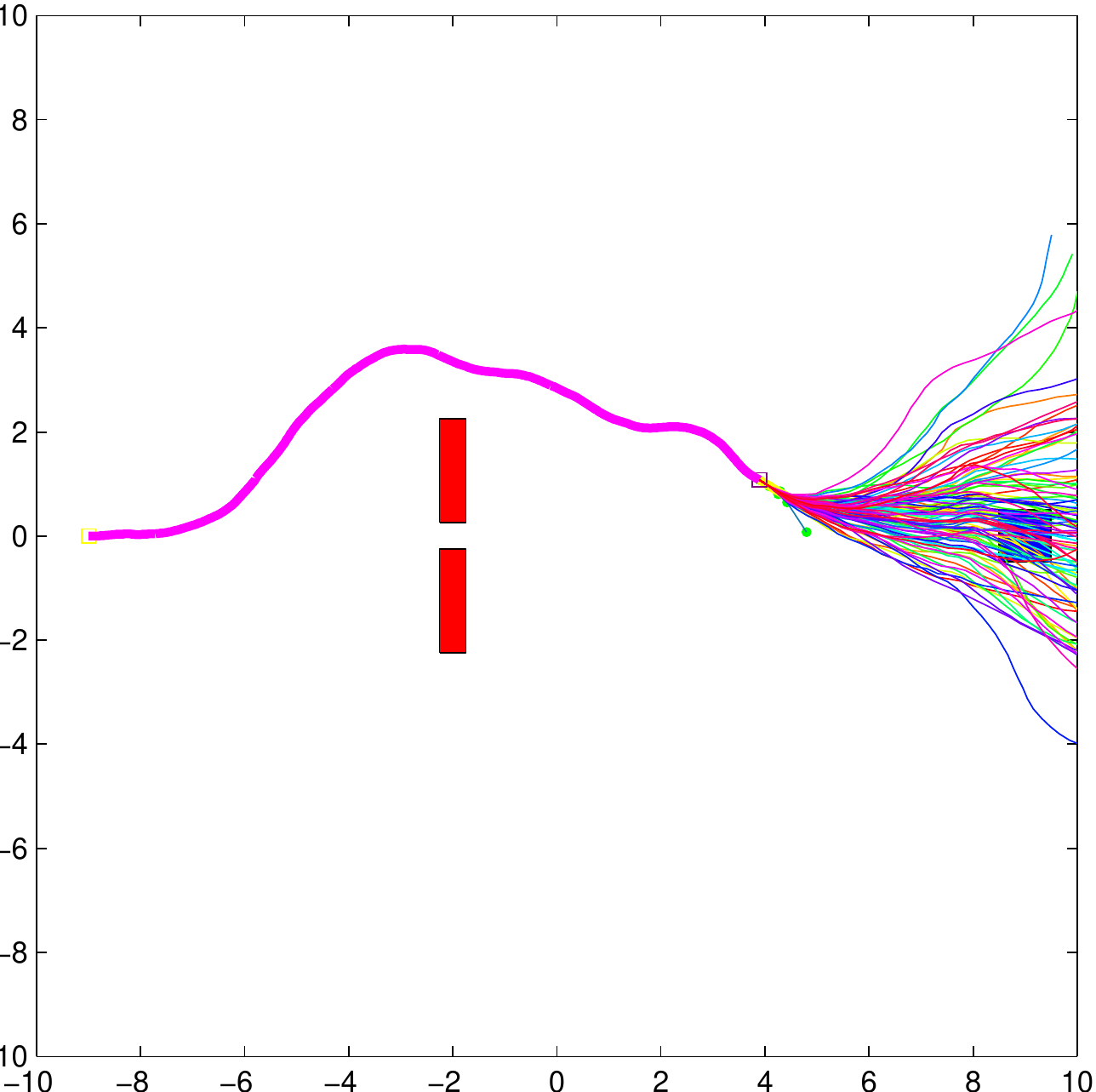}}\label{figure:pt1_alpha100_fr3}}
          }
	\caption{The trajectories computed by the \AlgRRTPI{} algorithm for stochastic optimal control of the kinematic car model under different levels of noise injected to the control channel: \subref{figure:pt1_alpha025_fr1}-\subref{figure:pt1_alpha025_fr3} is with $\alpha = 0.25$, \subref{figure:pt1_alpha050_fr1}-\subref{figure:pt1_alpha050_fr3} is with $\alpha = 0.50$, and \subref{figure:pt1_alpha100_fr1}-\subref{figure:pt1_alpha100_fr3} is with $\alpha = 1.0$.}

 \label{figure:sim_results_pt1}
\end{figure*}

The objective in this problem is to find trajectories for the vehicle in a square environment with a box-like obstacle having a single slit. 
The trajectories computed by the \AlgRRTPI{} algorithm at different stages are shown in
Figure~\ref{figure:sim_results_pt1}. The initial state is plotted as a yellow square and the goal region is shown in blue with magenta border (right-most). The computed path by the \AlgRRT{} algorithm following the unforced dynamics is shown in yellow. The locally sampled trajectories which are bundled around the yellow trajectory are shown in different colors. The trajectory of the vehicle due to execution of the control policy for some finite time horizon is shown in magenta.

To understand how the intensity of the noise level affects the patterns of the trajectories of the system, we run the algorithm and analyzed the situation for three different cases, $\alpha = 0.25$, $0.5$ and $1.0$ corresponding to low, medium and high intensity noise levels in the control channel. As shown in Figure~\ref{figure:sim_results_pt1} \subref{figure:pt1_alpha025_fr1}-\subref{figure:pt1_alpha025_fr3}, the \AlgRRTPI{} algorithm computes trajectories that pass through the slit most of the time when there is low intensity noise in the control channel. As a first step, the \AlgRRTPI{} algorithm computes a baseline trajectory using the \AlgRRT{} algorithm. The vertices and the edges of the tree computed by the \AlgRRT{} algorithm are shown in green and blue colors, respectively. During the simulations, it was observed that this baseline trajectory does not necessarily pass through the slit. The \AlgRRT{} algorithm sometimes returns a baseline trajectory that passes close by the upper or the lower sections of the obstacle due to both the noise which is observed in the dynamics and the randomized nature of the algorithm itself. The \AlgRRTPI{} algorithm then samples a bundle of trajectories around the baseline trajectory in order to compute the variation term for the new control input. The new control input is computed by summing up the baseline control policy returned by the \AlgRRT{} algorithm and the variation term, which is the weighted average of the contribution of each locally sampled trajectory. These weights are computed by using the cost information of each locally sampled trajectory. We observed that the distribution of the trajectories, which pass close to the upper or lower corners or through the slit, changes as the intensity of the noise increases. For higher intensity of the noise, the \AlgRRTPI{} algorithm computes trajectories which do not pass through the slit but rather pass close to the upper or lower corners. This change in the distribution of trajectories is shown in Figure~\ref{figure:sim_results_pt1} \subref{figure:pt1_alpha050_fr1}-\subref{figure:pt1_alpha050_fr3} for medium intensity noise and in Figure~\ref{figure:sim_results_pt1} \subref{figure:pt1_alpha100_fr1}-\subref{figure:pt1_alpha100_fr3} for high intensity noise.

\subsection{Example 2: Double-slit Obstacle}
Next, we consider a more challenging motion planning problem. In this case, there are two slits on the obstacle block and the length of the slits is longer than in the previous example. The longer length of the slits results in a higher probability of collision while traversing through the slit, which makes the motion planning problem more challenging.

\begin{figure*}[!htp]
\centering
	\mbox{  
    \subfigure[]{\scalebox{0.30}{\includegraphics[trim = 4.5cm 7.6cm 4.0cm 7.3cm, clip =
          true]{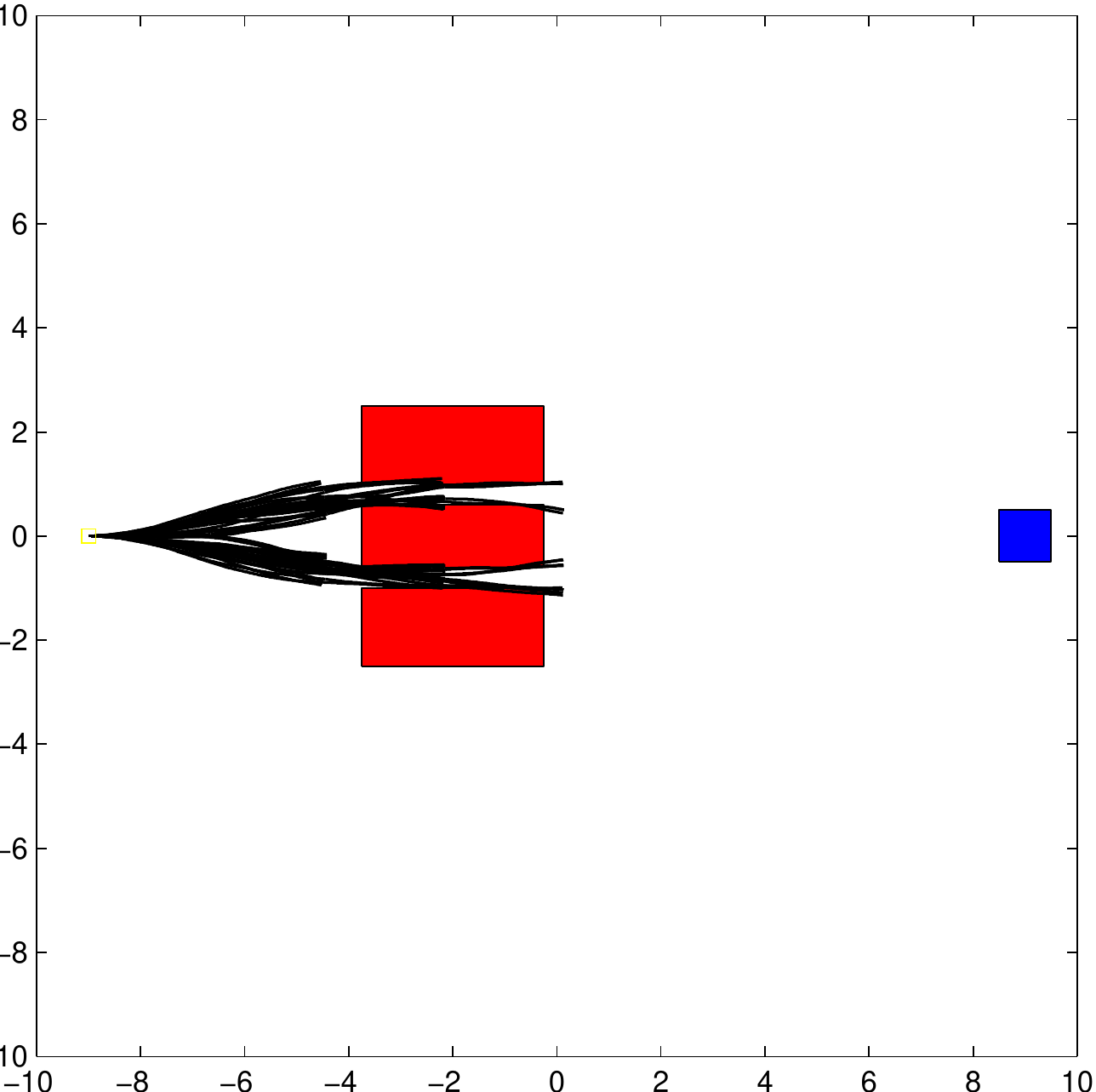}} \label{figure:pt2_alpha025_trials_rrt_fail}} 	
	\subfigure[]{\scalebox{0.30}{\includegraphics[trim = 4.5cm 7.6cm 4.0cm 7.3cm, clip =
          true]{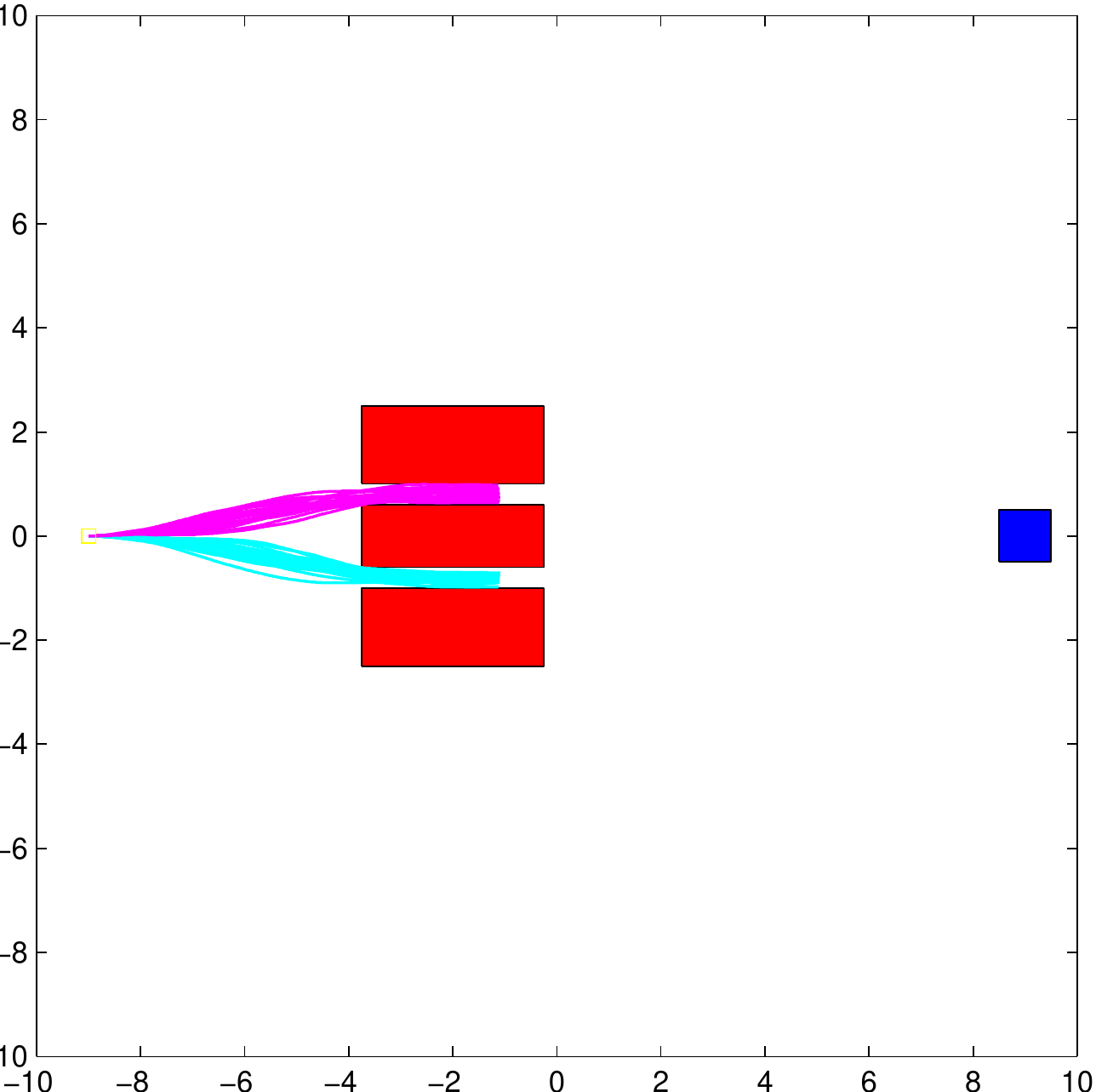}} \label{figure:pt2_alpha025_trials_rrt_success_fr2}}
	\subfigure[]{\scalebox{0.30}{\includegraphics[trim = 4.5cm 7.6cm 4.0cm 7.3cm, clip =
          true]{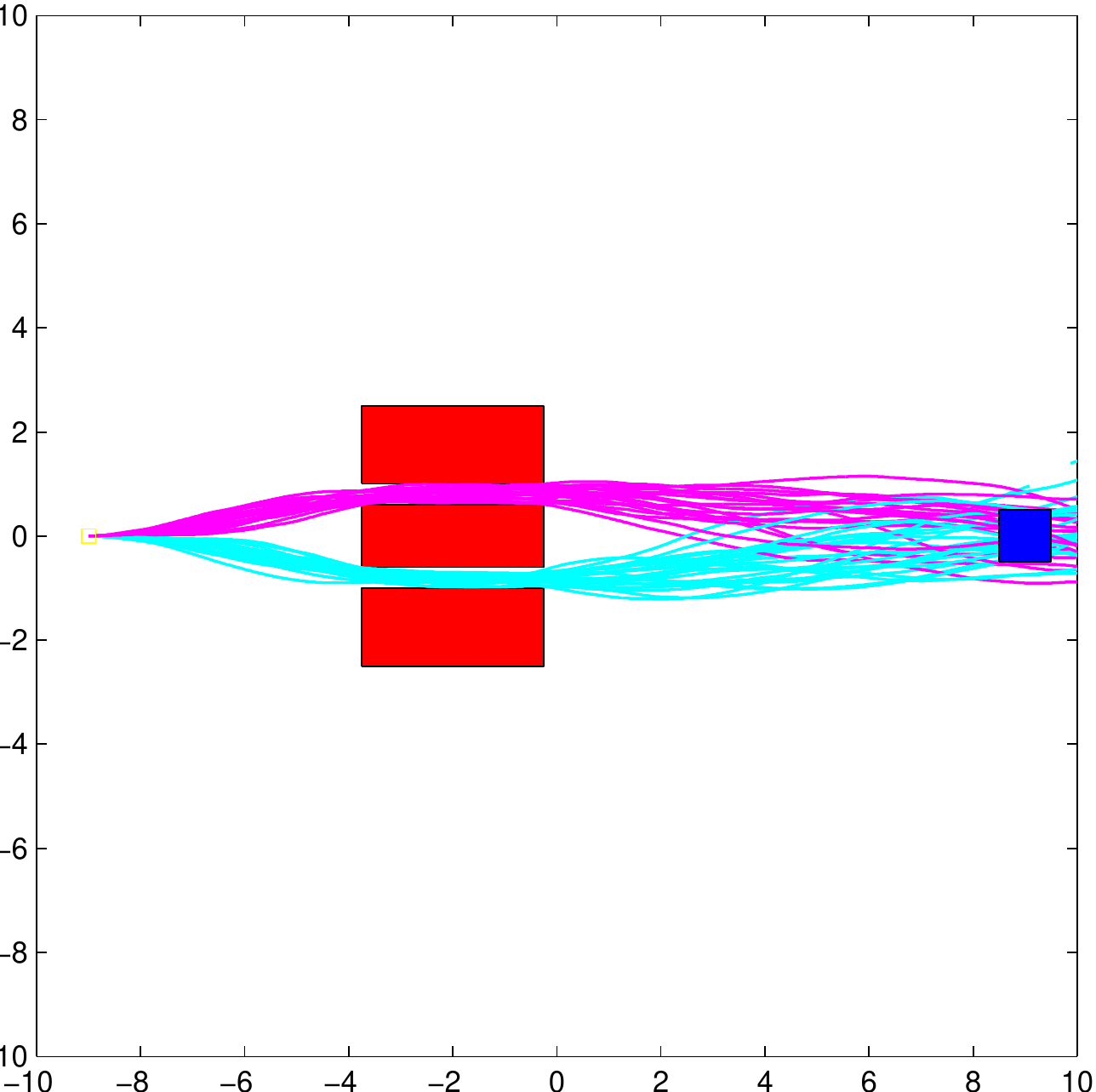}} \label{figure:pt2_alpha025_trials_rrt_success}}	
          }\\
	\mbox{  
    \subfigure[]{\scalebox{0.30}{\includegraphics[trim = 4.5cm 7.6cm 4.0cm 7.3cm, clip =
          true]{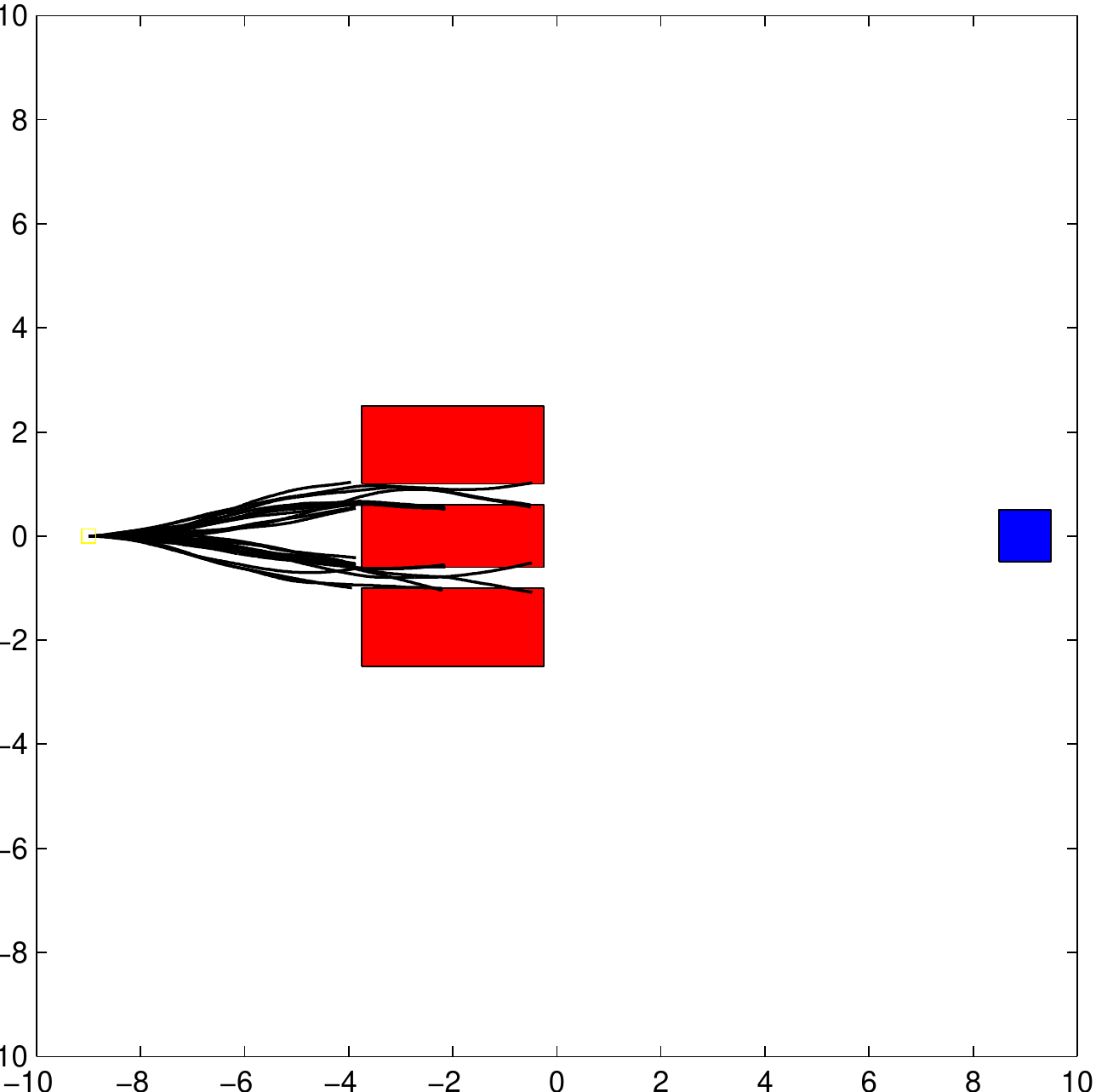}} \label{figure:pt2_alpha025_trials_rrt_pi_fail}} 	
	\subfigure[]{\scalebox{0.30}{\includegraphics[trim = 4.5cm 7.6cm 4.0cm 7.3cm, clip =
          true]{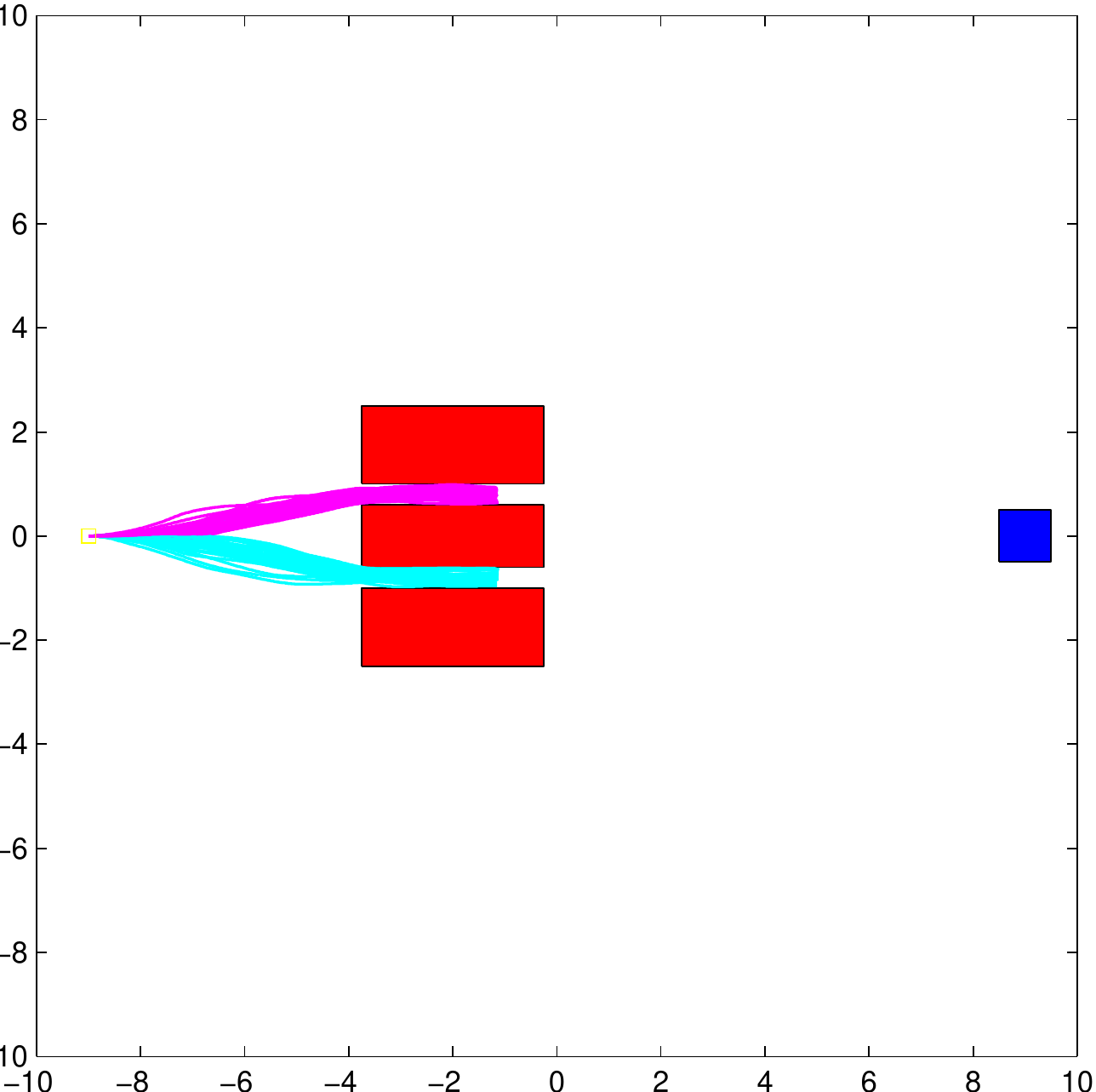}}
 \label{figure:pt2_alpha025_trials_rrt_pi_success_fr2}}	
    \subfigure[]{\scalebox{0.30}{\includegraphics[trim = 4.5cm 7.6cm 4.0cm 7.3cm, clip =
          true]{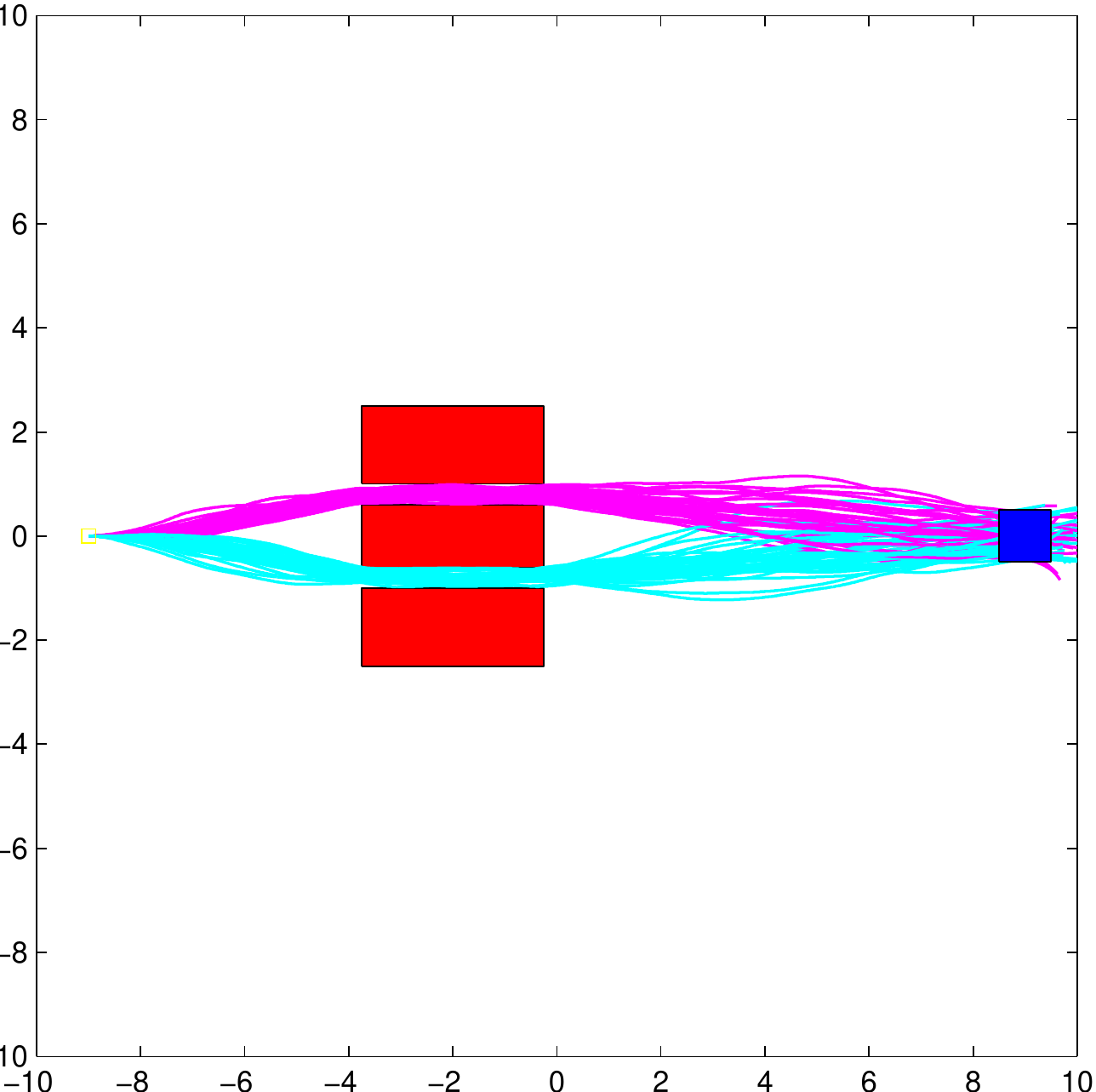}} \label{figure:pt2_alpha025_trials_rrt_pi_success}}			
          }

    	\caption{Distribution of trajectories for kinematic car model under low intensity of noise injected to the control channel ($\alpha = 0.25$) is shown in \subref{figure:pt2_alpha025_trials_rrt_fail}-\subref{figure:pt2_alpha025_trials_rrt_success} for the \AlgRRT{} algorithm, and in \subref{figure:pt2_alpha025_trials_rrt_pi_fail}-\subref{figure:pt2_alpha025_trials_rrt_pi_success} for the \AlgRRTPI{} algorithm. The trajectories which hit the obstacles are shown in \subref{figure:pt2_alpha025_trials_rrt_fail}, \subref{figure:pt2_alpha025_trials_rrt_pi_fail}. The collision-free trajectories at an intermediate stage are shown in \subref{figure:pt2_alpha025_trials_rrt_success_fr2}, \subref{figure:pt2_alpha025_trials_rrt_pi_success_fr2}, and at the final stage are shown in \subref{figure:pt2_alpha025_trials_rrt_success}, \subref{figure:pt2_alpha025_trials_rrt_pi_success}.}
    	
\label{figure:sim_results_alpha025_pt2_trials}
\end{figure*}

\begin{figure*}[!htp]
\centering
	\mbox{  
    \subfigure[]{\scalebox{0.30}{\includegraphics[trim = 4.5cm 7.6cm 4.0cm 7.3cm, clip =
          true]{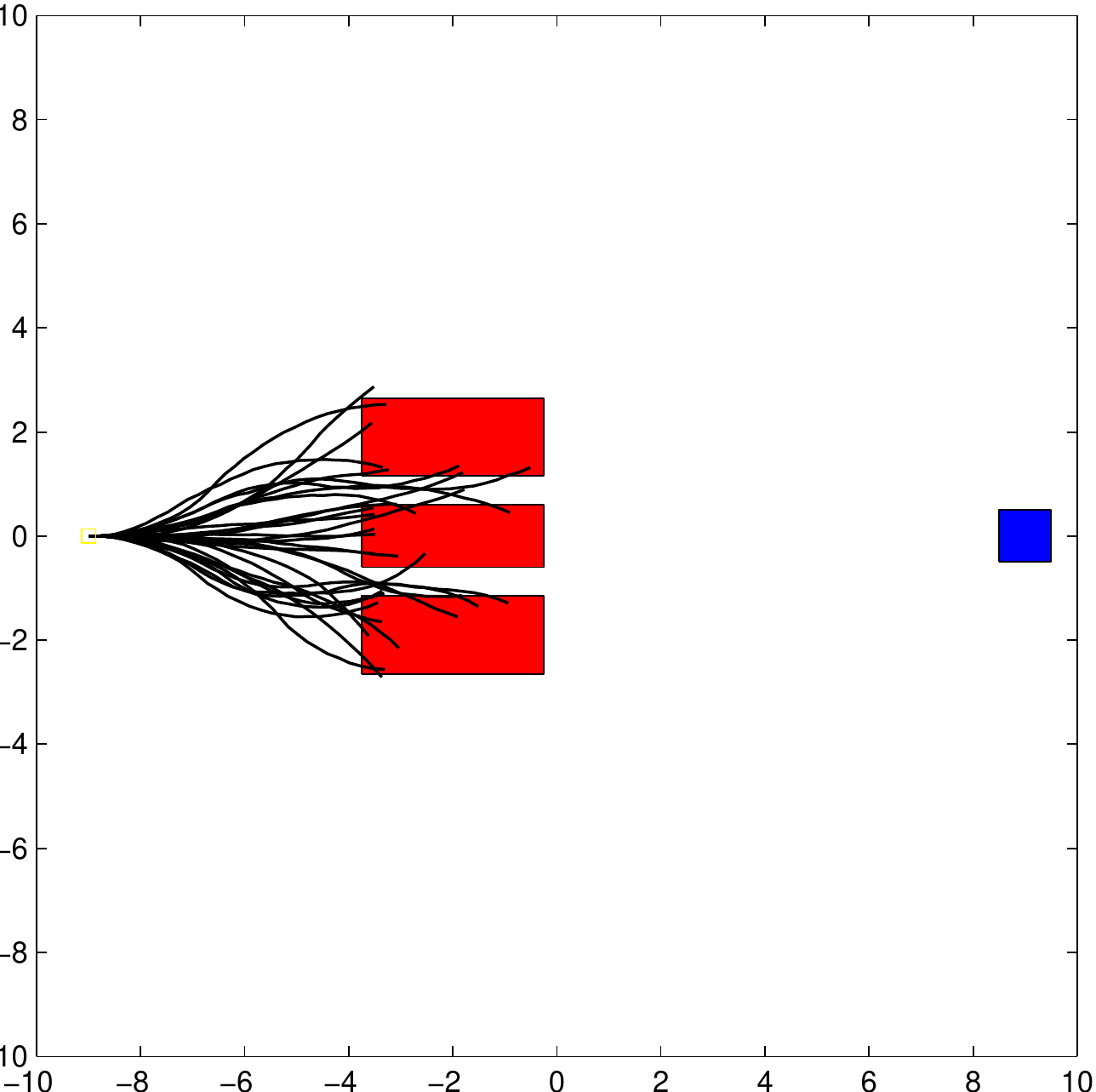}} \label{figure:pt2_alpha050_trials_rrt_fail}} 	
	\subfigure[]{\scalebox{0.30}{\includegraphics[trim = 4.5cm 7.6cm 4.0cm 7.3cm, clip =
          true]{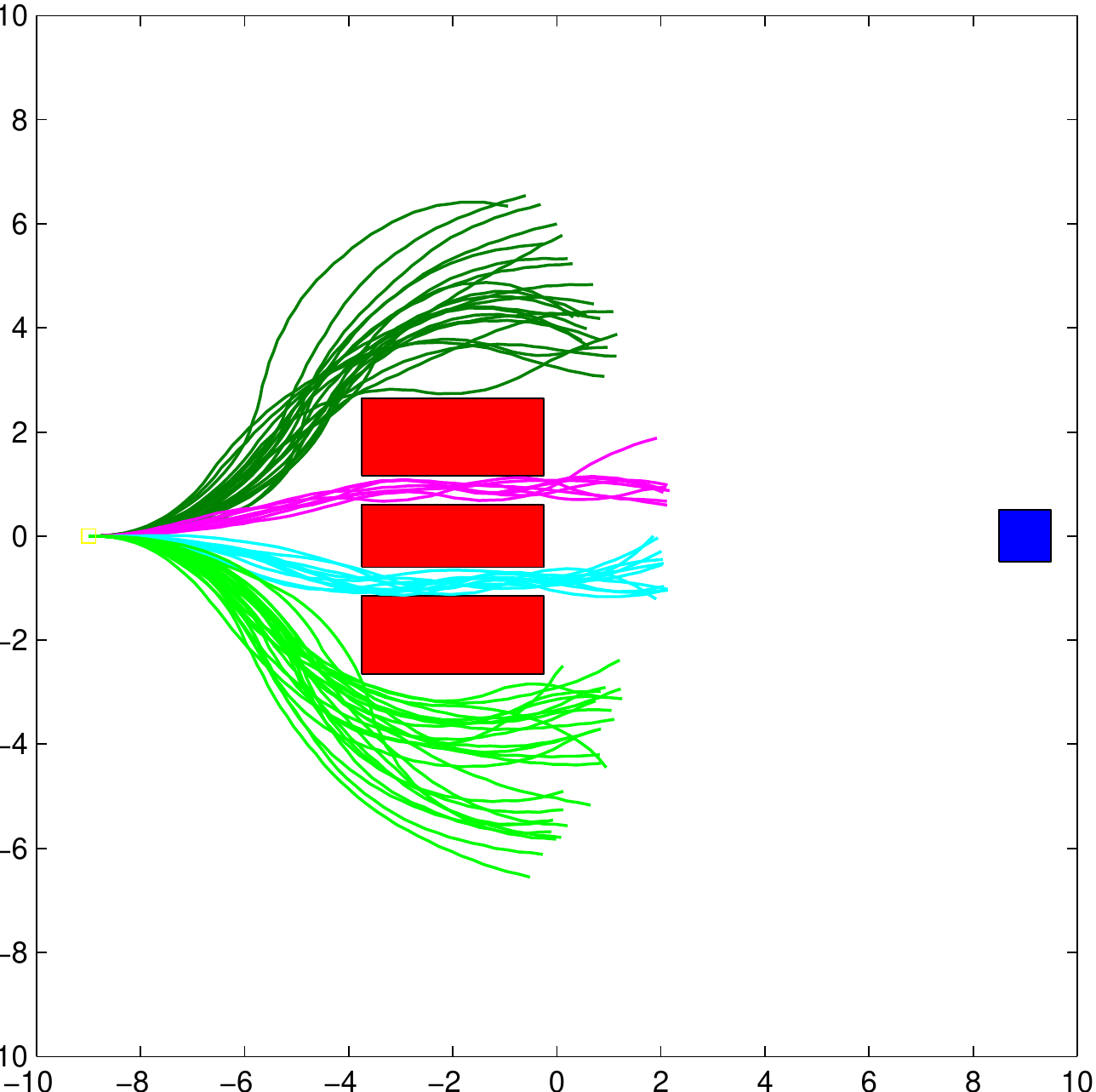}} \label{figure:pt2_alpha050_trials_rrt_success_fr2}}	
	\subfigure[]{\scalebox{0.30}{\includegraphics[trim = 4.5cm 7.6cm 4.0cm 7.3cm, clip =
          true]{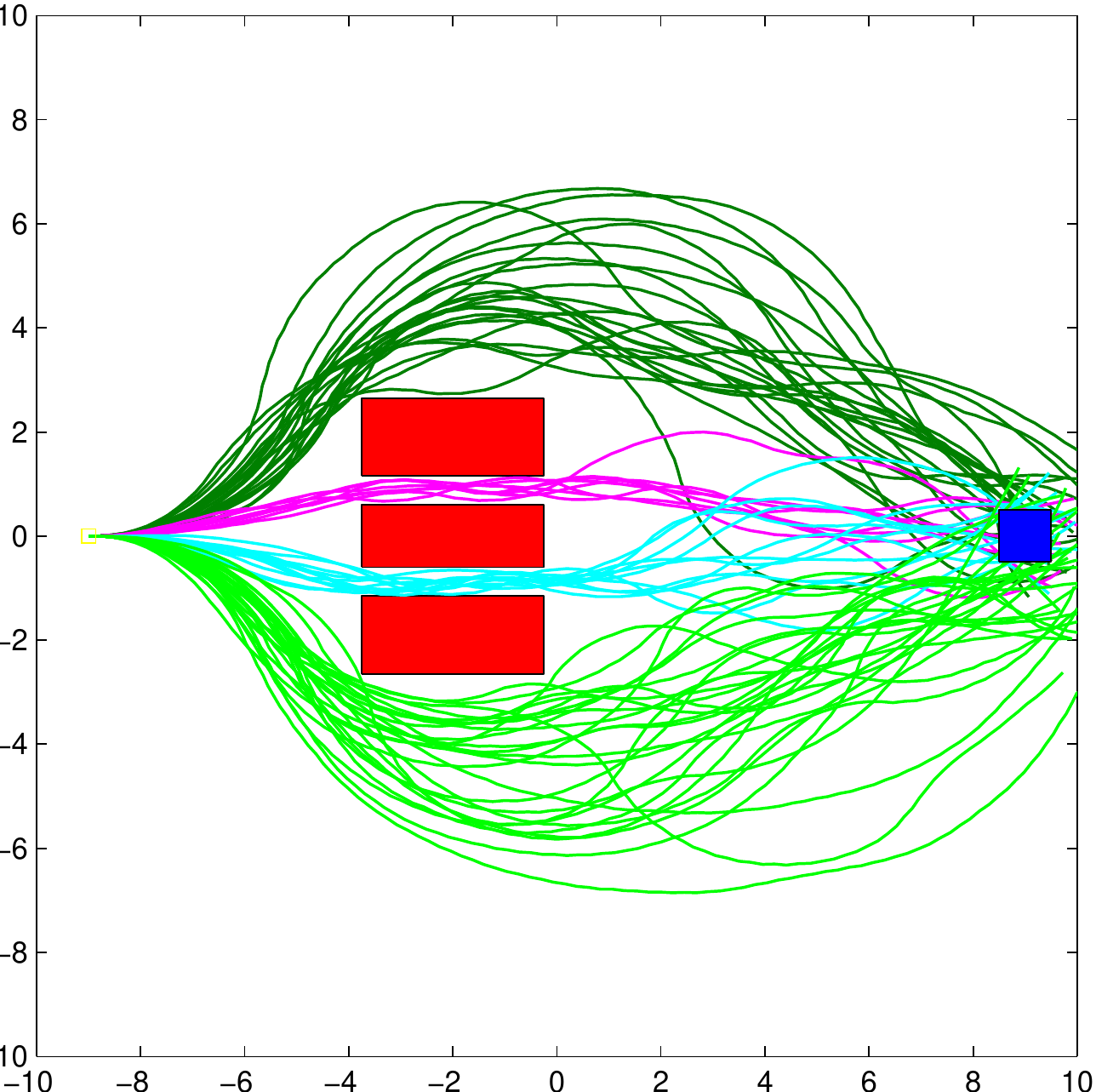}} \label{figure:pt2_alpha050_trials_rrt_success}}	
          }\\
	\mbox{  
    \subfigure[]{\scalebox{0.30}{\includegraphics[trim = 4.5cm 7.6cm 4.0cm 7.3cm, clip =
          true]{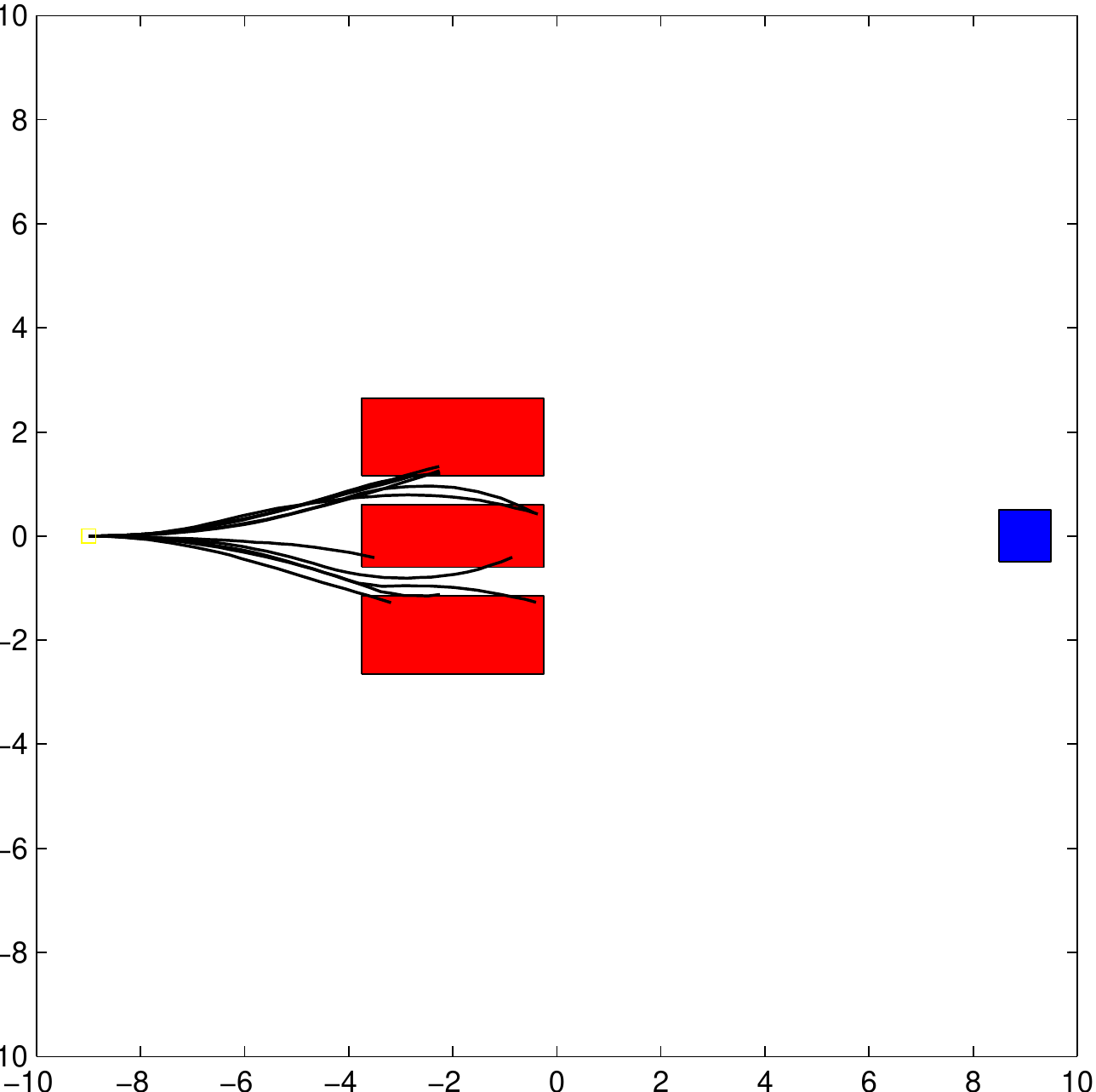}} \label{figure:pt2_alpha050_trials_rrt_pi_fail}} 	
	\subfigure[]{\scalebox{0.30}{\includegraphics[trim = 4.5cm 7.6cm 4.0cm 7.3cm, clip =
          true]{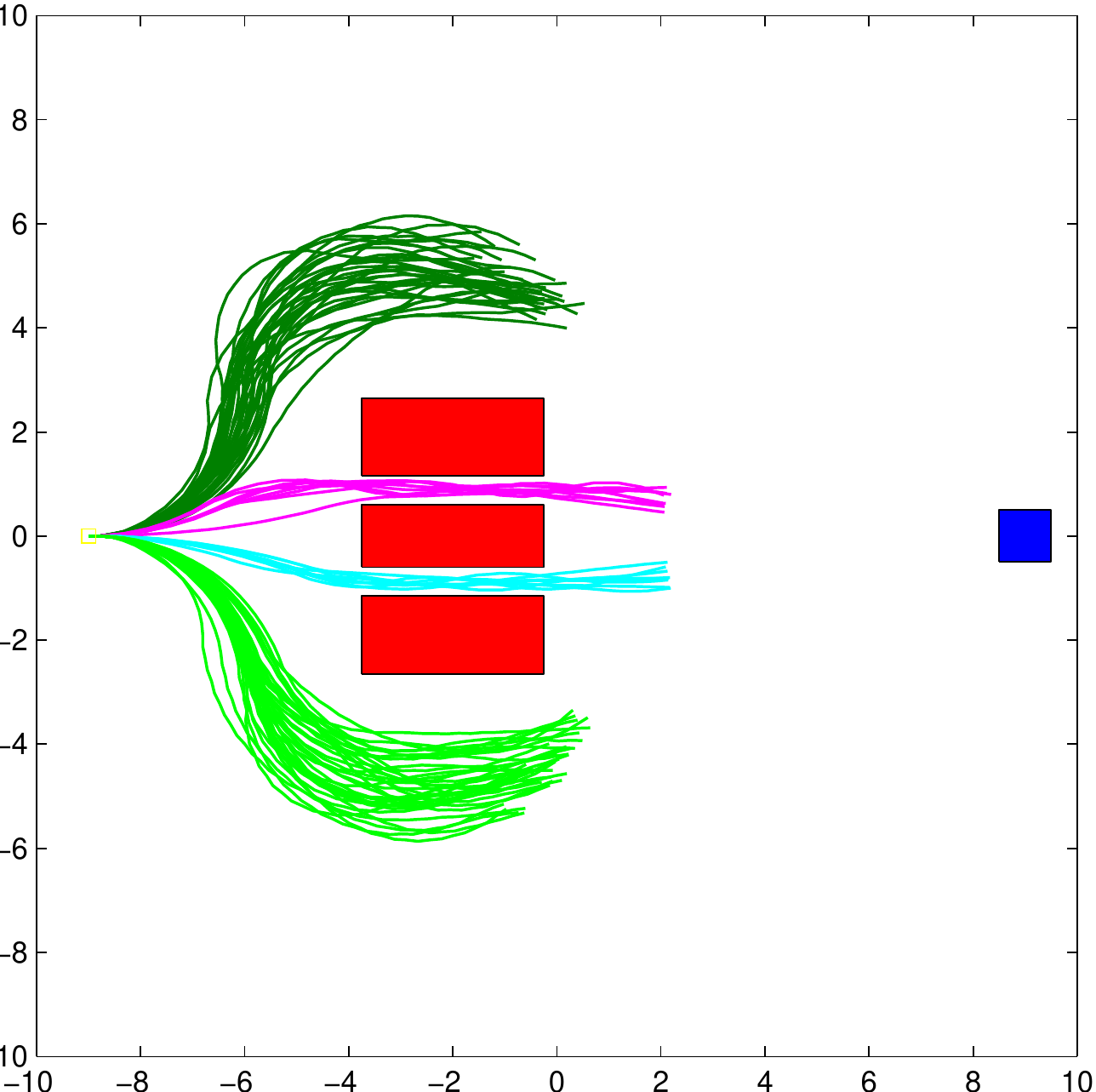}} \label{figure:pt2_alpha050_trials_rrt_pi_success_fr2}}		
	\subfigure[]{\scalebox{0.30}{\includegraphics[trim = 4.5cm 7.6cm 4.0cm 7.3cm, clip =
          true]{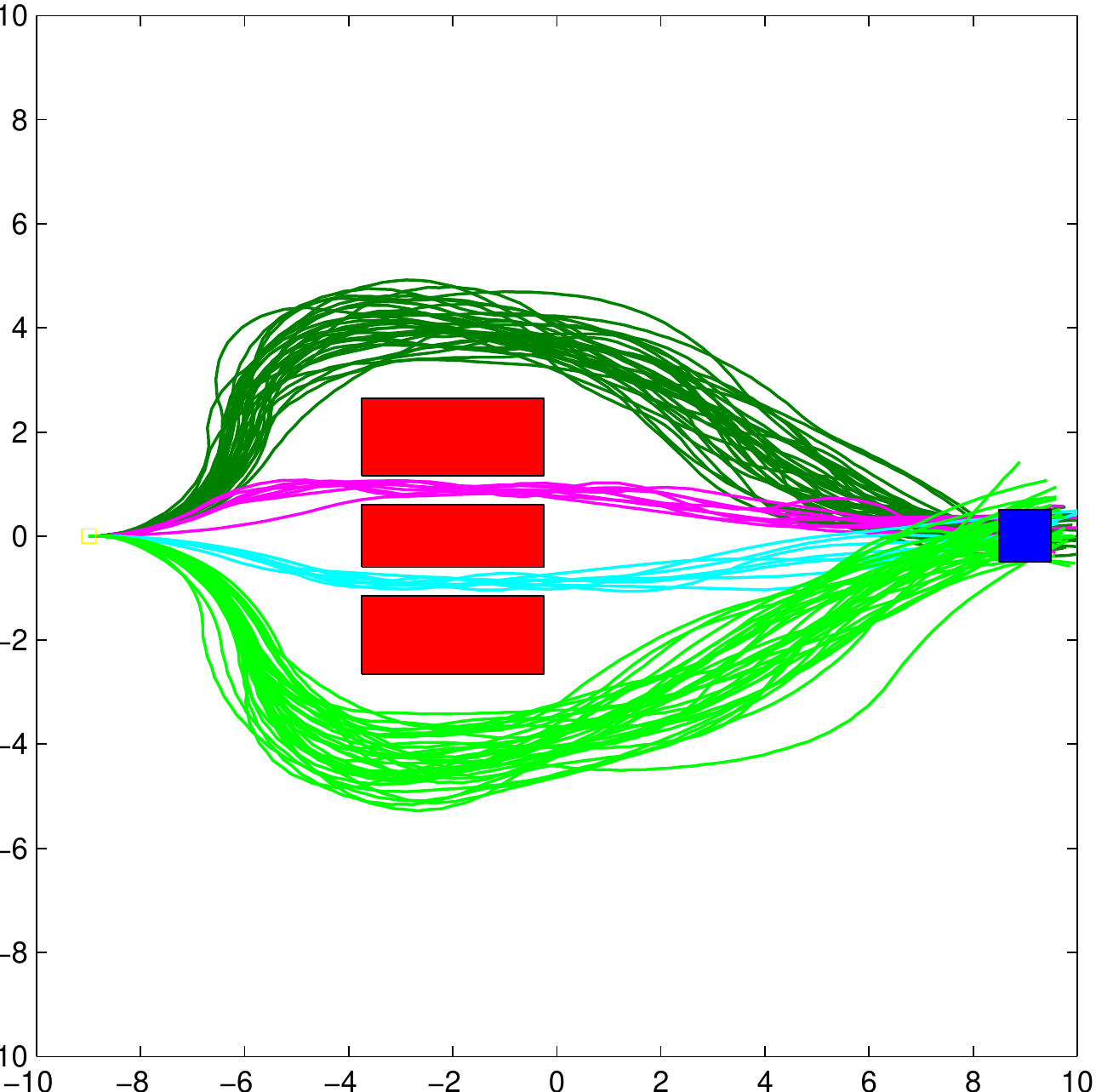}} \label{figure:pt2_alpha050_trials_rrt_pi_success}}			
          }

    	\caption{Distribution of trajectories for kinematic car model under low intensity of noise injected to the control channel ($\alpha = 0.50$) is shown in \subref{figure:pt2_alpha050_trials_rrt_fail}-\subref{figure:pt2_alpha050_trials_rrt_success} for the \AlgRRT{} algorithm, and in \subref{figure:pt2_alpha050_trials_rrt_pi_fail}-\subref{figure:pt2_alpha050_trials_rrt_pi_success} for the \AlgRRTPI{} algorithm. The trajectories which hit the obstacles are shown in \subref{figure:pt2_alpha050_trials_rrt_fail}, \subref{figure:pt2_alpha050_trials_rrt_pi_fail}. The collision-free trajectories at an intermediate stage are shown in \subref{figure:pt2_alpha050_trials_rrt_success_fr2}, \subref{figure:pt2_alpha050_trials_rrt_pi_success_fr2}, and at the final stage are shown in \subref{figure:pt2_alpha050_trials_rrt_success}, \subref{figure:pt2_alpha050_trials_rrt_pi_success}.}
    	
\label{figure:sim_results_alpha050_pt2_trials}
\end{figure*}

\begin{figure*}[!htp]
\centering
	\mbox{  
    \subfigure[]{\scalebox{0.30}{\includegraphics[trim = 4.5cm 7.6cm 4.0cm 7.3cm, clip =
          true]{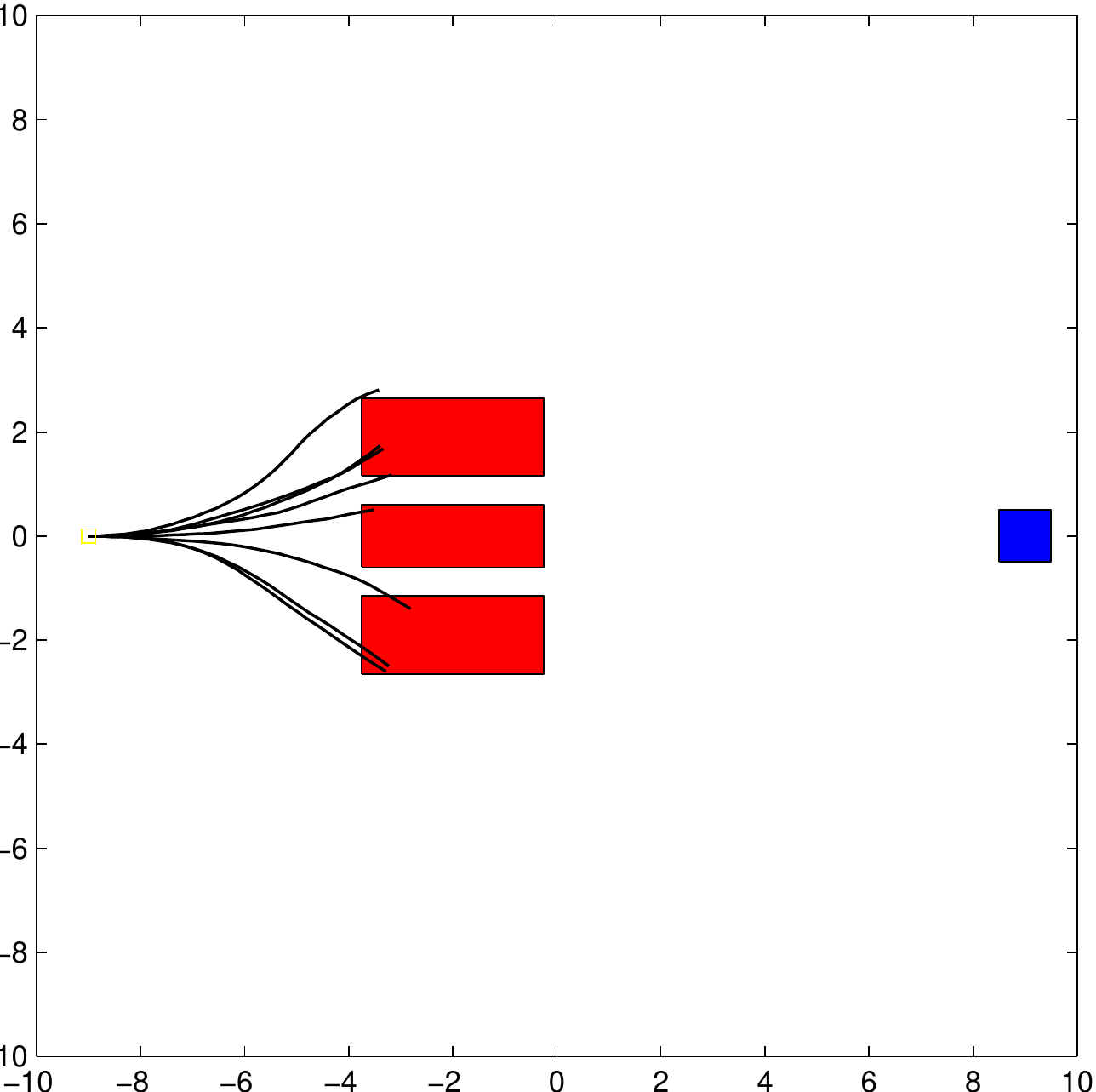}} \label{figure:pt2_alpha100_trials_rrt_fail}} 	
	\subfigure[]{\scalebox{0.30}{\includegraphics[trim = 4.5cm 7.6cm 4.0cm 7.3cm, clip =
          true]{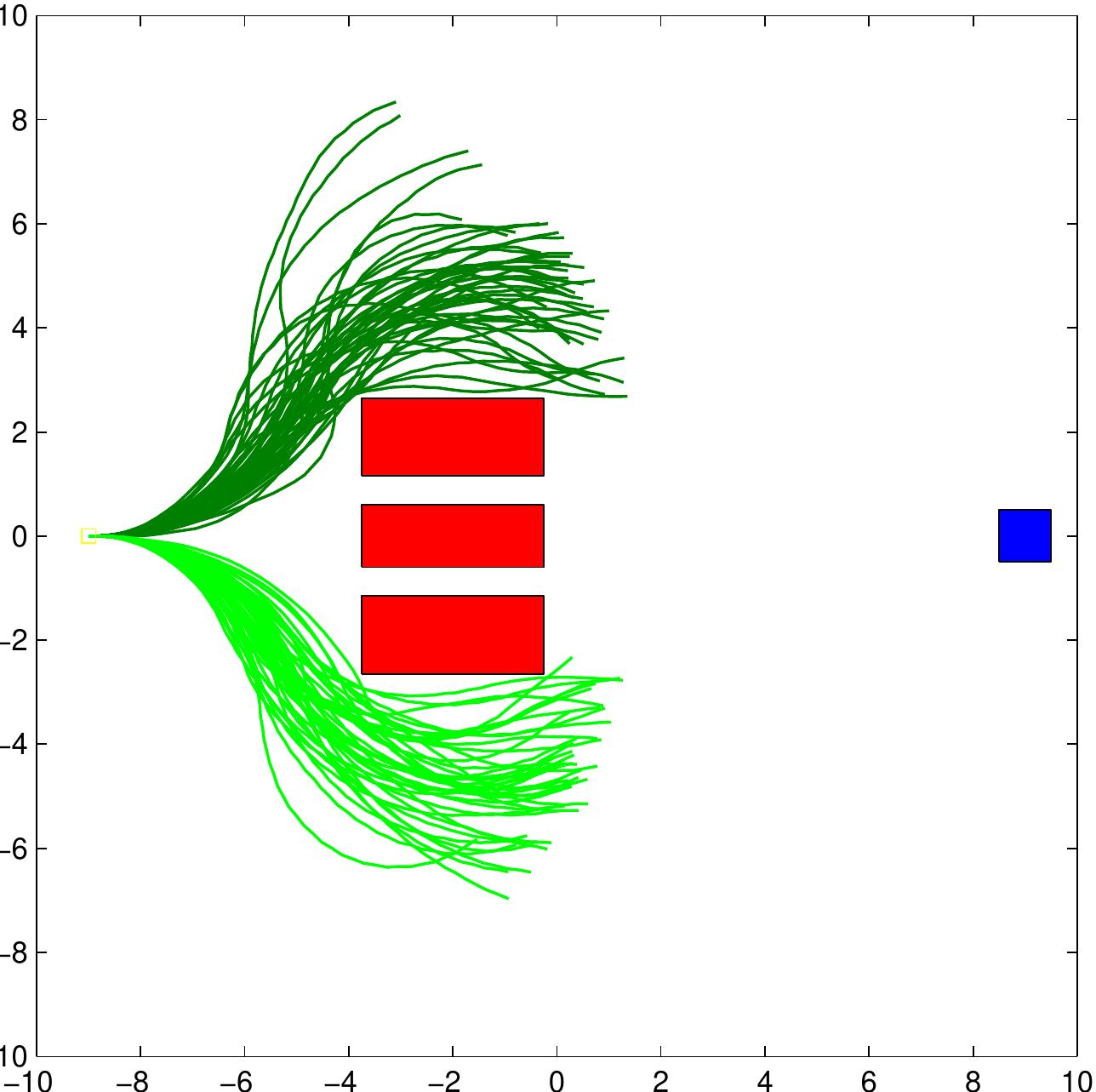}} \label{figure:pt2_alpha100_trials_rrt_success_fr2}}	
	\subfigure[]{\scalebox{0.30}{\includegraphics[trim = 4.5cm 7.6cm 4.0cm 7.3cm, clip =
          true]{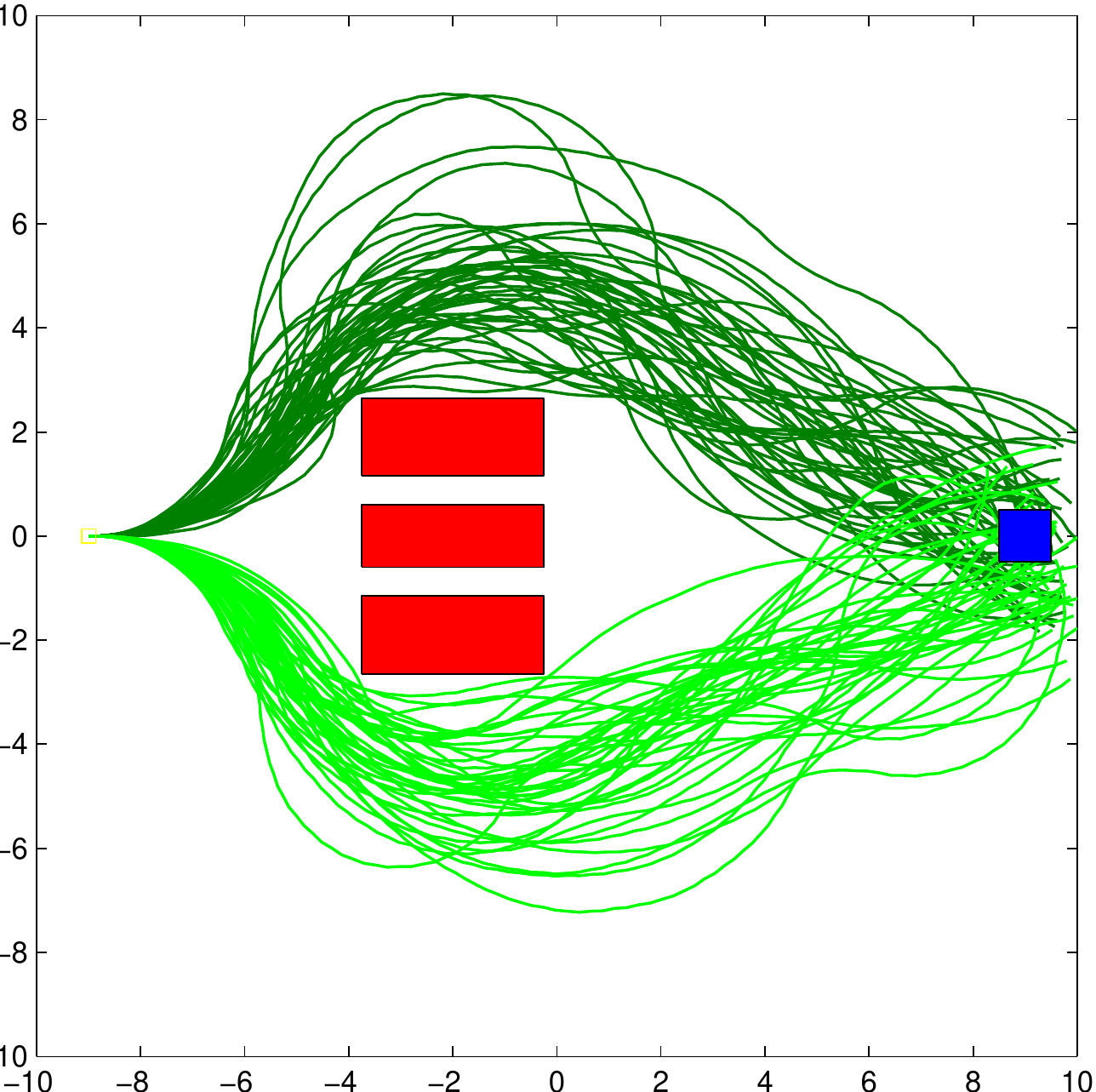}} \label{figure:pt2_alpha100_trials_rrt_success}}	
          }\\
	\mbox{  
    \subfigure[]{\scalebox{0.30}{\includegraphics[trim = 4.5cm 7.6cm 4.0cm 7.3cm, clip =
          true]{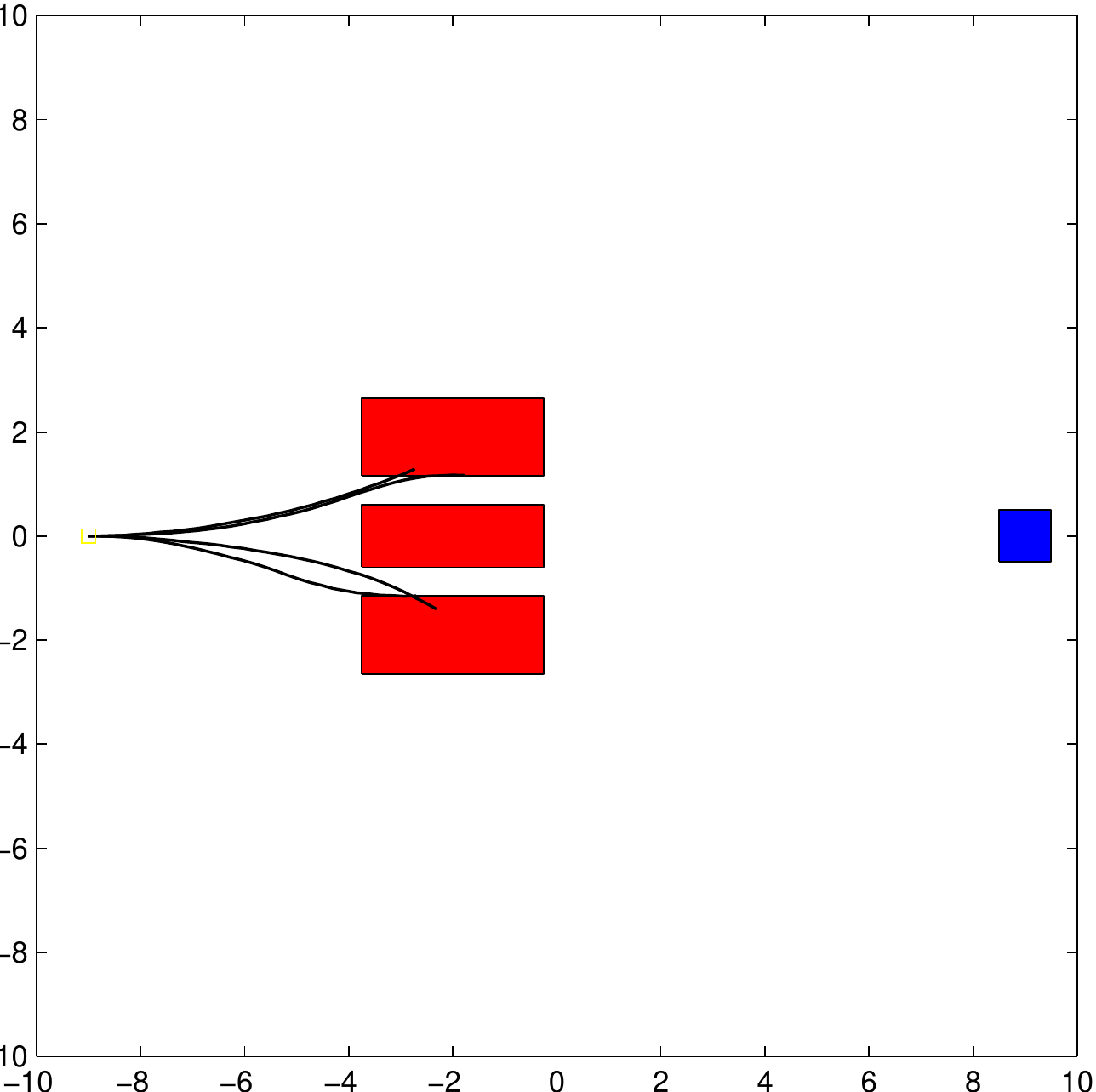}} \label{figure:pt2_alpha100_trials_rrt_pi_fail}} 	
	\subfigure[]{\scalebox{0.30}{\includegraphics[trim = 4.5cm 7.6cm 4.0cm 7.3cm, clip =
          true]{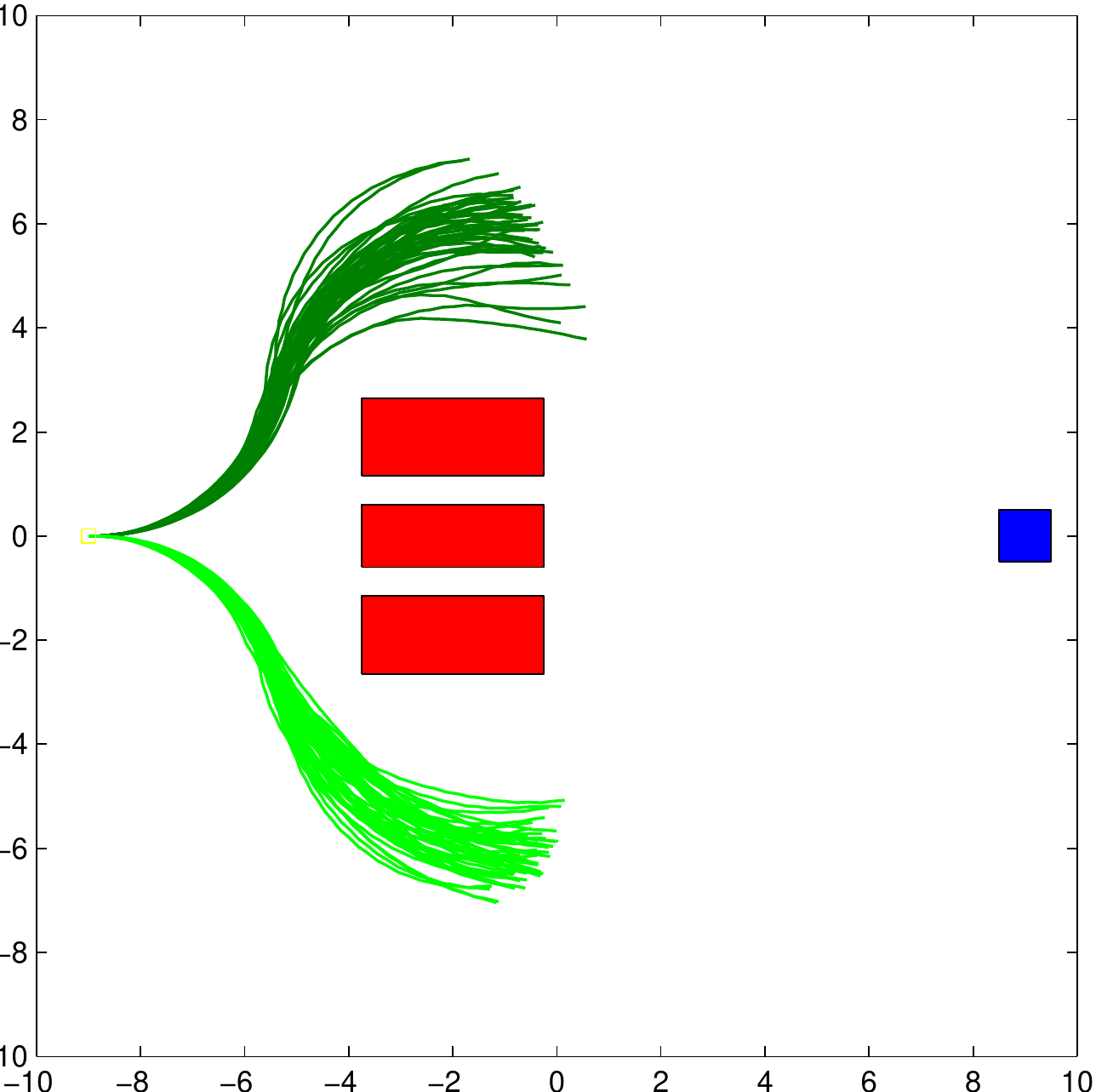}} \label{figure:pt2_alpha100_trials_rrt_pi_success_fr2}}	
	\subfigure[]{\scalebox{0.30}{\includegraphics[trim = 4.5cm 7.6cm 4.0cm 7.3cm, clip =
          true]{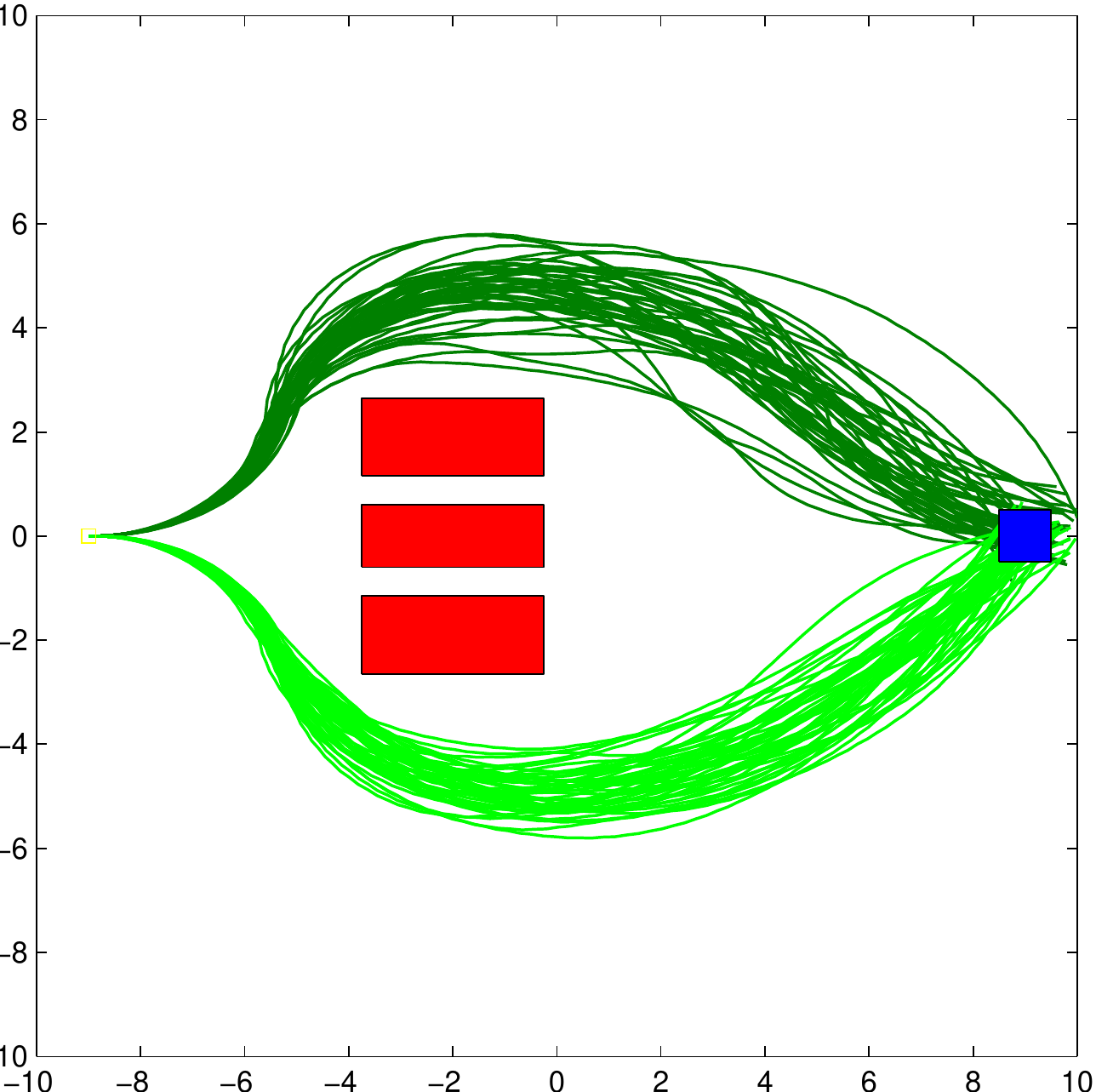}} \label{figure:pt2_alpha100_trials_rrt_pi_success}}			
          }

    	\caption{Distribution of trajectories for kinematic car model under low intensity of noise injected to the control channel ($\alpha = 1.0$) is shown in \subref{figure:pt2_alpha100_trials_rrt_fail}-\subref{figure:pt2_alpha100_trials_rrt_success} for the \AlgRRT{} algorithm, and in \subref{figure:pt2_alpha100_trials_rrt_pi_fail}-\subref{figure:pt2_alpha100_trials_rrt_pi_success} for the \AlgRRTPI{} algorithm. The trajectories which hit the obstacles are shown in \subref{figure:pt2_alpha100_trials_rrt_fail}, \subref{figure:pt2_alpha100_trials_rrt_pi_fail}. The collision-free trajectories at an intermediate stage are shown in \subref{figure:pt2_alpha100_trials_rrt_success_fr2}, \subref{figure:pt2_alpha100_trials_rrt_pi_success_fr2}, and at the final stage are shown in \subref{figure:pt2_alpha100_trials_rrt_success}, \subref{figure:pt2_alpha100_trials_rrt_pi_success}.}
    	
\label{figure:sim_results_alpha100_pt2_trials}
\end{figure*}

A study was performed in order to compare the performance of the \AlgRRTPI{} algorithm with the \AlgRRT{} algorithm. No variation term in the control input was computed for the \AlgRRT{} algorithm, and it was simply executed in a receding horizon fashion. All algorithms were run for 6000 iterations to find a baseline trajectory. The results over 100 trials are shown in Figures~\ref{figure:sim_results_alpha025_pt2_trials}, \ref{figure:sim_results_alpha050_pt2_trials} and \ref{figure:sim_results_alpha100_pt2_trials}. The trajectories that result in collision are plotted in Figure~\ref{figure:sim_results_alpha025_pt2_trials} \subref{figure:pt2_alpha025_trials_rrt_fail}, \subref{figure:pt2_alpha025_trials_rrt_pi_fail} for the low noise level, Figure~\ref{figure:sim_results_alpha050_pt2_trials} \subref{figure:pt2_alpha050_trials_rrt_fail}, \subref{figure:pt2_alpha050_trials_rrt_pi_fail} for the medium noise level, and Figure~\ref{figure:sim_results_alpha100_pt2_trials} \subref{figure:pt2_alpha100_trials_rrt_fail}, \subref{figure:pt2_alpha100_trials_rrt_pi_fail} for the high noise level for the \AlgRRT{} and \AlgRRTPI{} algorithms, respectively. Also, the distribution of collision-free trajectories is plotted in Figure~\ref{figure:sim_results_alpha025_pt2_trials} \subref{figure:pt2_alpha025_trials_rrt_success}, \subref{figure:pt2_alpha025_trials_rrt_pi_success} for the low noise level, Figure~\ref{figure:sim_results_alpha050_pt2_trials} \subref{figure:pt2_alpha050_trials_rrt_success}, \subref{figure:pt2_alpha050_trials_rrt_pi_success} for the medium noise level, and Figure~\ref{figure:sim_results_alpha100_pt2_trials} \subref{figure:pt2_alpha100_trials_rrt_success}, \subref{figure:pt2_alpha100_trials_rrt_pi_success} for the high noise level for the \AlgRRT{} and \AlgRRTPI{} algorithms, respectively. The distribution of trajectories and the number of trajectories which result in a collision are summarized in Table I. Under the `Success' column, the rows of the table contain the number of collision-free trajectories which pass through the bottom corner, bottom slit, top slit and top corner of the block. As shown in Table~\ref{table:monte_carlo_double_slit}, the \AlgRRTPI{} computes safer control policies which reduce the risk of having a collision. On the other hand, both the \AlgRRT{} and the \AlgRRTPI{} compute trajectories that are almost equally distributed over both slits.

In summary, it was observed that the behaviors of both algorithms are similar for the case with high noise level. As the noise level decreases, most of the failed cases, not surprisingly, occur when the algorithms try to compute a path that passes through the slits. Our simulation results demonstrate that the \AlgRRTPI{} algorithm tends to compute trajectories that have larger clearance from obstacles and hence outperforms the standard \AlgRRT{} algorithm, resulting in a smaller failure rate.


\begin{table}[h]
\caption{Monte-Carlo Results for Double-Slit Obstacle}\label{table:monte_carlo_double_slit}
\centering
\begin{tabular}{l|cccc|c||cccc|c||cccc|c|}
\cline{2-16}
 & \multicolumn{5}{|c||}{$\alpha = 0.25$}         & \multicolumn{5}{|c||}{$\alpha = 0.50$}          & \multicolumn{5}{|c|}{$\alpha = 1.00$}            \\ \hline
\multicolumn{1}{|l||}{Algorithm}         & \multicolumn{4}{|c|}{Success} & Fail  & \multicolumn{4}{|c|}{Success} & Fail & \multicolumn{4}{|c|}{Success} & Fail \\ \hline
\multicolumn{1}{|l||}{RRT}     & 0     & 24     & 20    & 0    & 56    & 23     & 8    & 11    & 27    & 31   & 48     & 0     & 0     & 44   & 8    \\ \hline
\multicolumn{1}{|l||}{\AlgRRTPI} & 0     & 44     & 45    & 0    & 11    & 35     & 9    & 8     & 37    & 11   & 47     & 0     & 0     & 49   & 4    \\ \hline
\end{tabular}
\end{table}

\section{Conclusion}
In this paper, the \AlgRRTPI{} algorithm is proposed in order to solve a class of stochastic optimal control problems. The proposed approach makes a novel connection between incremental sampling-based algorithms and path integral control. The work in this paper can be extended in several directions. First, a parallel version of the algorithm can be implemented by sampling local trajectories or computing several initial trajectories simultaneously. Second, since there exist many variants of the standard \AlgRRT{} algorithm, one can implement different sampling-based algorithms to compute initial trajectories and incorporate them within the path integral framework. For example, the \AlgRRTstar{}~\cite{karaman2010optimal,karaman2011sampling} and the \AlgRRTsharp{} algorithms~\cite{arslan2013use}, which are both asymptotically optimal, can be used to compute bundles of good initial trajectories in a single pass; however, such an algorithm would require more elaborate computations for implementing the steering function, e.g., backward integration of a stochastic differential equation. This is part of ongoing work.

\bibliographystyle{plain}
\bibliography{arslan.arxiv.final}

\clearpage
\addtocmark[2]{Author Index} 
\renewcommand{\indexname}{Author Index}
\printindex
\clearpage
\addtocmark[2]{Subject Index} 
\markboth{Subject Index}{Subject Index}
\renewcommand{\indexname}{Subject Index}
\end{document}

%% file: commands_integrated.tex
\usepackage[english]{babel}
\usepackage[latin1]{inputenc}

\usepackage{amsmath,amsfonts,amssymb}
\usepackage{xfrac}
\usepackage{color}
\usepackage[english]{babel}
\usepackage{graphicx}
\usepackage{placeins}
\usepackage{multirow}

\usepackage{comment}

\DeclareMathOperator*{\E}{\mathbb{E}}
\DeclareMathOperator{\Tr}{tr}

\newcommand{\xinitial}{x_{\mathrm{i}}}	
\newcommand{\yinitial}{y_{\mathrm{i}}}	
\newcommand{\thetainitial}{\theta_{\mathrm{i}}}	

\newcommand{\xfinal}{x_{\mathrm{f}}}	
\newcommand{\yfinal}{y_{\mathrm{f}}}	
\newcommand{\thetafinal}{\theta_{\mathrm{f}}}	

\newcommand{\AlgRRTPI}{PI-RRT}

\newcommand{\scX}{\mathcal{X}} 		
\newcommand{\scZ}{\mathcal{Z}} 		

\newcommand{\scXGoal}{\scX_{\mathrm{goal}}} 
\newcommand{\scZGoal}{\scZ_{\mathrm{goal}}} 
\newcommand{\cTGoal}{\ensuremath{T_{\mathrm{goal}}}}

\newcommand{\lTInit}{\ensuremath{t_\mathrm{init}}}

\newcommand{\lZInit}{z_{\mathrm{init}}}	
\newcommand{\lXInit}{x_{\mathrm{init}}}	

\newcommand{\dtau}{\mathrm{d}\tau}

\newcommand{\tf}{\mathrm{t_{f}}}
\newcommand{\ti}{\mathrm{t_{i}}}

\newcommand{\dif}[1]{\mathrm{d}{#1}}

\definecolor{gray}{RGB}{190,190,190}
\definecolor{darkgreen}{rgb}{0,0.5,0} 
\definecolor{fullred}{rgb}{0.85,.0,.1} 
\definecolor{brown}{rgb}{0.65,0.16,0.16}

\newcommand{\AlgRRT}{\ensuremath{{\mathrm{RRT}}}}

\newcommand{\AlgRRTstar}{\ensuremath{{\mathrm{RRT}^*}}}
\newcommand{\AlgRRTsharp}{\ensuremath{{\mathrm{RRT}^{\tiny \#}}}}

\newcommand{\naturals}{\mathbb{N}} 	

\newcommand{\Z}{\mathcal{Z}} 		
\newcommand{\Zfree}{\Z_{\mathrm{free}}} 

\newcommand{\zrand}{z_{\mathrm{rand}}}	

\newcommand{\graph}[1]{\mathcal{#1}}		

\newcommand{\True}{{\tt True}}
\newcommand{\False}{{\tt False}}

\newcommand{\PrcSample}{\mathtt{Sample}} 
\newcommand{\PrcExtend}{\mathtt{Extend}} 
\newcommand{\PrcNearest}{\mathtt{Nearest}} 
\newcommand{\PrcSteer}{\mathtt{Steer}}   
\newcommand{\PrcObstacleFree}{\mathtt{ObstacleFree}} 

\newcommand{\hideMaterial}[1]{}
\newcommand{\hideoldproofofnegresult}[1]{ }

%% file: algorithms/rrt.tex
\IncMargin{1em}
\begin{algorithm}[htp]
    \small
    \DontPrintSemicolon
    \SetKwInOut{Input}{input}
    \SetKwInOut{Output}{output}
    \SetKwBlock{NoBegin}{}{end}


    \SetKwFunction{fRRT}{\AlgRRT}
    \SetKwFunction{fSample}{Sample}
    \SetKwFunction{fExtend}{Extend}
    \SetKwFunction{fReduceInconsistency}{ReduceInconsistency}
    \SetKwFunction{fParent}{parent}


	\SetKwData{vZNew}{$\mathbf{z}_{\mathrm{new}}$}   
	\SetKwData{vSigmaNew}{$\sigma_{\mathrm{new}}$}	
	\SetKwData{vUNew}{$\mathbf{u}_{\mathrm{new}}$}
	\SetKwData{vZRand}{$\mathbf{z}_{\mathrm{rand}}$}
	\SetKwData{vZ}{$\mathbf{z}$}    
    \SetKwData{vZInit}{$\mathbf{z}_{\mathrm{init}}$}
    \SetKwData{vCZGoal}{$\mathcal{Z}_{\mathrm{goal}}$}
    \SetKwData{vCZ}{$\mathcal{Z}$}
    \SetKwData{vV}{$V$}
    \SetKwData{vE}{$E$}
    \SetKwData{vEPrime}{$E^{\prime}$}
    \SetKwData{vXInit}{$x_{\mathrm{init}}$}
    \SetKwData{vI}{$i$}
    \SetKwData{vXRand}{$x_{\mathrm{rand}}$}
    \SetKwData{vT}{$\mathcal{T}$}
    \SetKwData{vG}{$\mathcal{G}$}
    \SetKwData{vCXGoal}{$\mathcal{X}_{\mathrm{goal}}$}
    \SetKwData{vCX}{$\mathcal{X}$}
    \SetKwData{vX}{$x$}

	\SetFuncSty{textbf}
    \fRRT{\vZInit, \vCZGoal, \vCZ}
    \SetFuncSty{texttt}
    \NoBegin
    {
        $\vV \leftarrow \{\vZInit\}$;
        $\vE \leftarrow \emptyset$;

        $\vG \leftarrow (\vV,\vE)$;

        \For{$\vI = 1$ to $N$ \label{line:rrtsharp_itbegin}}
        {
            $\vZRand \leftarrow \fSample(\vI)$;

            $\vG \leftarrow \fExtend(\vG, \vZRand)$;


        }

%

        \Return{$\vG$}
    }

\caption{Body of the \AlgRRT{} Algorithm\label{alg:rrt}}
\end{algorithm}
\DecMargin{1em}

%% file: algorithms/extend_rrt.tex
\IncMargin{1em}
\begin{algorithm}[htp]
	\small

    \DontPrintSemicolon
    \SetKwInOut{Input}{input}
    \SetKwInOut{Output}{output}
    \SetKwBlock{NoBegin}{}{end}


	\SetKwFunction{fExtend}{Extend}      
    \SetKwFunction{fNearest}{Nearest}
    \SetKwFunction{fSteer}{Steer}
    \SetKwFunction{fObstacleFree}{ObstacleFree}
    \SetKwFunction{fUpdateState}{UpdateQueue}
    \SetKwFunction{fNear}{Near}
    \SetKwFunction{fG}{g}
    \SetKwFunction{fC}{c}
    \SetKwFunction{fLMC}{lmc}
    \SetKwFunction{fParent}{parent}
    \SetKwFunction{fInitState}{Initialize}



	\SetKwData{vZNew}{$\mathbf{z}_{\mathrm{new}}$}    
	\SetKwData{vSigmaNew}{{\mbox{\boldmath$\sigma$}}$_{\mathrm{new}}$}	
	\SetKwData{vUNew}{$\mathbf{u}_{\mathrm{new}}$}
	\SetKwData{vZNearest}{$\mathbf{z}_{\mathrm{nearest}}$}
	\SetKwData{vZ}{$\mathbf{z}$}
    \SetKwData{vT}{$\mathcal{T}$}
    \SetKwData{vG}{$\mathcal{G}$}
    \SetKwData{vTPrime}{$\mathcal{T}^{\prime}$}
    \SetKwData{vGPrime}{$\mathcal{G}^{\prime}$}
    \SetKwData{vX}{$x$}
    \SetKwData{vV}{$V$}
    \SetKwData{vE}{$E$}
    \SetKwData{vEPrime}{$E^{\prime}$}
    \SetKwData{vXNearest}{$x_{\mathrm{nearest}}$}
    \SetKwData{vXNew}{$x_{\mathrm{new}}$}
    \SetKwData{vCXNear}{$\mathcal{X}_{\mathrm{near}}$}
    \SetKwData{vXNear}{$x_{\mathrm{near}}$}



    \SetFuncSty{textbf}
	\fExtend{\vG,\vZ}
    \SetFuncSty{texttt}
    \NoBegin
    {
		
		$(\vV,\vE) \leftarrow \vG$;	
        
        $\vZNearest \leftarrow \fNearest(\vG,\vZ)$;
         
        $(\vZNew, \vSigmaNew, \vUNew) \leftarrow \fSteer(\vZNearest, \vZ)$; 
        
		\If{$\fObstacleFree(\vSigmaNew)$}
        {
			$\vV \leftarrow \vV \cup \{ \vZNew \}$;

			$\vE \leftarrow \vE \cup \{ (\vZNearest, \vZNew) \}$;
        }

        \Return{$\vGPrime \leftarrow (\vV,\vE)$}
    }

%
%

\caption{${\tt Extend}$ Procedure for \AlgRRT{} Algorithm\label{alg:extend_rrt}} 
\end{algorithm}
\DecMargin{1em} 

%% file: algorithms/rrt_pi.tex
\IncMargin{1em}
\begin{algorithm}[htp]
    \small
    \DontPrintSemicolon
    \SetKwInOut{Input}{input}
    \SetKwInOut{Output}{output}
    \SetKwBlock{NoBegin}{}{end}


    \SetKwFunction{fRRTPI}{\AlgRRTPI}
    \SetKwFunction{fRRT}{\AlgRRT}
    \SetKwFunction{fExtractPath}{ExtractPath}
    \SetKwFunction{fExecute}{Execute}
    \SetKwFunction{fSamplePath}{SamplePath}
    \SetKwFunction{fMeasureState}{MeasureState}
    \SetKwFunction{fComputeVariation}{ComputeVariation}
	\SetKwFunction{fJ}{J}
	\SetKwFunction{fJPrime}{J$^{\prime}$}
    
    \SetKwFunction{fSample}{Sample}
    \SetKwFunction{fExtend}{Extend}
    \SetKwFunction{fReduceInconsistency}{ReduceInconsistency}
    \SetKwFunction{fParent}{parent}


    
    \SetKwData{vDeltaU}{$\delta \mathbf{u}$}
    \SetKwData{vTInit}{$t_{\mathrm{init}}$}
    \SetKwData{vTi}{$t_{\mathrm{i}}$}
    \SetKwData{vTf}{$t_{\mathrm{f}}$}
    \SetKwData{vXi}{$\mathbf{x}_{\mathrm{i}}$}
    \SetKwData{vK}{$k$}
    \SetKwData{vZi}{$\mathbf{z}_{\mathrm{i}}$}
    \SetKwData{vV}{$V$}
    \SetKwData{vE}{$E$}
    \SetKwData{vEPrime}{$E^{\prime}$}
    \SetKwData{vXInit}{$x_{\mathrm{init}}$}
    \SetKwData{vXInit}{$\mathbf{x}_{\mathrm{init}}$}
    \SetKwData{vZInit}{$\mathbf{z}_{\mathrm{init}}$}
    \SetKwData{vI}{$i$}
    \SetKwData{vXRand}{$\mathbf{x}_{\mathrm{rand}}$}
    \SetKwData{vZRand}{$\mathbf{z}_{\mathrm{rand}}$}
    \SetKwData{vT}{$\mathcal{T}$}
    \SetKwData{vG}{$\mathcal{G}$}
    \SetKwData{vCXGoal}{$\mathcal{X}_{\mathrm{goal}}$}
    \SetKwData{vTGoal}{$T_{\mathrm{goal}}$}
    \SetKwData{vCZGoal}{$\mathcal{Z}_{\mathrm{goal}}$}
    \SetKwData{vCX}{$\mathcal{X}$}
    \SetKwData{vCZ}{$\mathcal{Z}$}
    \SetKwData{vX}{$\mathbf{x}$}
    \SetKwData{vZ}{$\mathbf{z}$}
    \SetKwData{vSigmaK}{{\mbox{\boldmath$\sigma$}}$_{\mathrm{k}}$}	
    \SetKwData{vSigma}{{\mbox{\boldmath$\sigma$}}}	
	\SetKwData{vU}{$\mathbf{u}$}
    \SetKwData{vUk}{$\mathbf{u}_{\mathrm{k}}$}
    \SetKwData{vWk}{$\mathbf{w}_{\mathrm{k}}$}
	\SetKwData{vFu}{$f_{\mathrm{u}}$}
    \SetKwData{vV}{$V$}
    \SetKwData{vE}{$E$}
    \SetKwData{vEPrime}{$E^{\prime}$}
    \SetKwData{vXInit}{$x_{\mathrm{init}}$}
    \SetKwData{vI}{$i$}
    \SetKwData{vXRand}{$x_{\mathrm{rand}}$}
    \SetKwData{vT}{$\mathcal{T}$}
    \SetKwData{vG}{$\mathcal{G}$}
    \SetKwData{vCXGoal}{$\mathcal{X}_{\mathrm{goal}}$}
    \SetKwData{vCX}{$\mathcal{X}$}
    \SetKwData{vX}{$\mathbf{x}$}

	\SetFuncSty{textbf}
    \fRRTPI{\vZInit, \vCZGoal, \vCZ}
    \SetFuncSty{texttt}
    \NoBegin
    {
    	$(\vTi, \vXi) \leftarrow \vZInit$;
    	$(\vTf, \vCXGoal) \leftarrow \vCZGoal$;
    	
    	$\vZi \leftarrow \vZInit$;
    	
    	\While{$\vTi < \vTf$}
    	{

    		$\vG \leftarrow \fRRT{\vZi, \vCZGoal, \vCZ}$;
    	
    		$( \vSigma_{\AlgRRT}, \vU_{\AlgRRT}) \leftarrow \fExtractPath(\vG)$;
    	
    		
    		$\vDeltaU_{[\vTi, \vTf]} \leftarrow \fComputeVariation(\vU_{\AlgRRT}, M)$;
       		
       		

    		$\vU \leftarrow \vU_{\AlgRRT} + \vDeltaU$;
    	 
        	$\fExecute( \vU_{[ \ti, \ti + \tau ]})$;
        	
        	$\vXi \leftarrow \fMeasureState(\vTi + \tau)$;
        	$\vTi \leftarrow \vTi + \tau$;
        	
        	$\vZi \leftarrow (\vXi,\vTi)$;

		}
    }

\caption{Body of the \AlgRRTPI{} Algorithm\label{alg:rrt_pi}}
\end{algorithm}
\DecMargin{1em}